\documentclass[10pt,twocolumn,letterpaper]{article}

\usepackage{wacv}
\usepackage{times}
\usepackage{epsfig}
\usepackage{graphicx}
\usepackage{amsmath}
\usepackage{amssymb}
\usepackage{booktabs}
\usepackage{multirow}
\usepackage{csquotes}
\usepackage{color}
\usepackage[table,xcdraw,dvipsnames]{xcolor}
\usepackage{bbding}
\usepackage{authblk}
\usepackage{graphbox}

\usepackage[accsupp]{axessibility}  
\usepackage{ulem}
\usepackage{soul}

\def\wacvPaperID{555} 

\wacvalgorithmstrack   

\wacvfinalcopy 

\ifwacvfinal
\usepackage[breaklinks=true,bookmarks=false]{hyperref}
\else
\usepackage[pagebackref=true,breaklinks=true,colorlinks,bookmarks=false]{hyperref}
\fi

\begin{document}

\title{I See-Through You: A Framework for Removing Foreground Occlusion in Both Sparse and Dense Light Field Images}

\newcommand\CoAuthorMark{\footnotemark[\arabic{footnote}]}

\author{Jiwan Hur\thanks{Equal contribution.}, Jae Young Lee\protect\CoAuthorMark, Jaehyun Choi, and Junmo Kim \\
School of Electrical Engineering, KAIST, South Korea \\
{\tt\small \{jiwan.hur, mcneato, chlwogus, junmo.kim\}@kaist.ac.kr}
}

\maketitle
\thispagestyle{empty}

\begin{abstract}
Light field (LF) camera captures rich information from a scene.
Using the information, the LF de-occlusion (LF-DeOcc) task aims to reconstruct the occlusion-free center view image.
Existing LF-DeOcc studies mainly focus on the sparsely sampled (sparse) LF images where most of the occluded regions are visible in other views due to the large disparity.
In this paper, we expand LF-DeOcc in more challenging datasets, densely sampled (dense) LF images, which are taken by a micro-lens-based portable LF camera.
Due to the small disparity ranges of dense LF images, most of the background regions are invisible in any view.
To apply LF-DeOcc in both LF datasets, we propose a framework, ISTY, which is defined and divided into three roles: (1) extract LF features, (2) define the occlusion, and (3) inpaint occluded regions.
By dividing the framework into three specialized components according to the roles, the development and analysis can be easier.
Furthermore, an explainable intermediate representation, an occlusion mask, can be obtained in the proposed framework.
The occlusion mask is useful for comprehensive analysis of the model and other applications by manipulating the mask.
In experiments, qualitative and quantitative results show that the proposed framework outperforms state-of-the-art LF-DeOcc methods in both sparse and dense LF datasets.

\end{abstract}

\section{Introduction}

\begin{figure}[!t]
\begin{center}
\includegraphics[width=\linewidth]{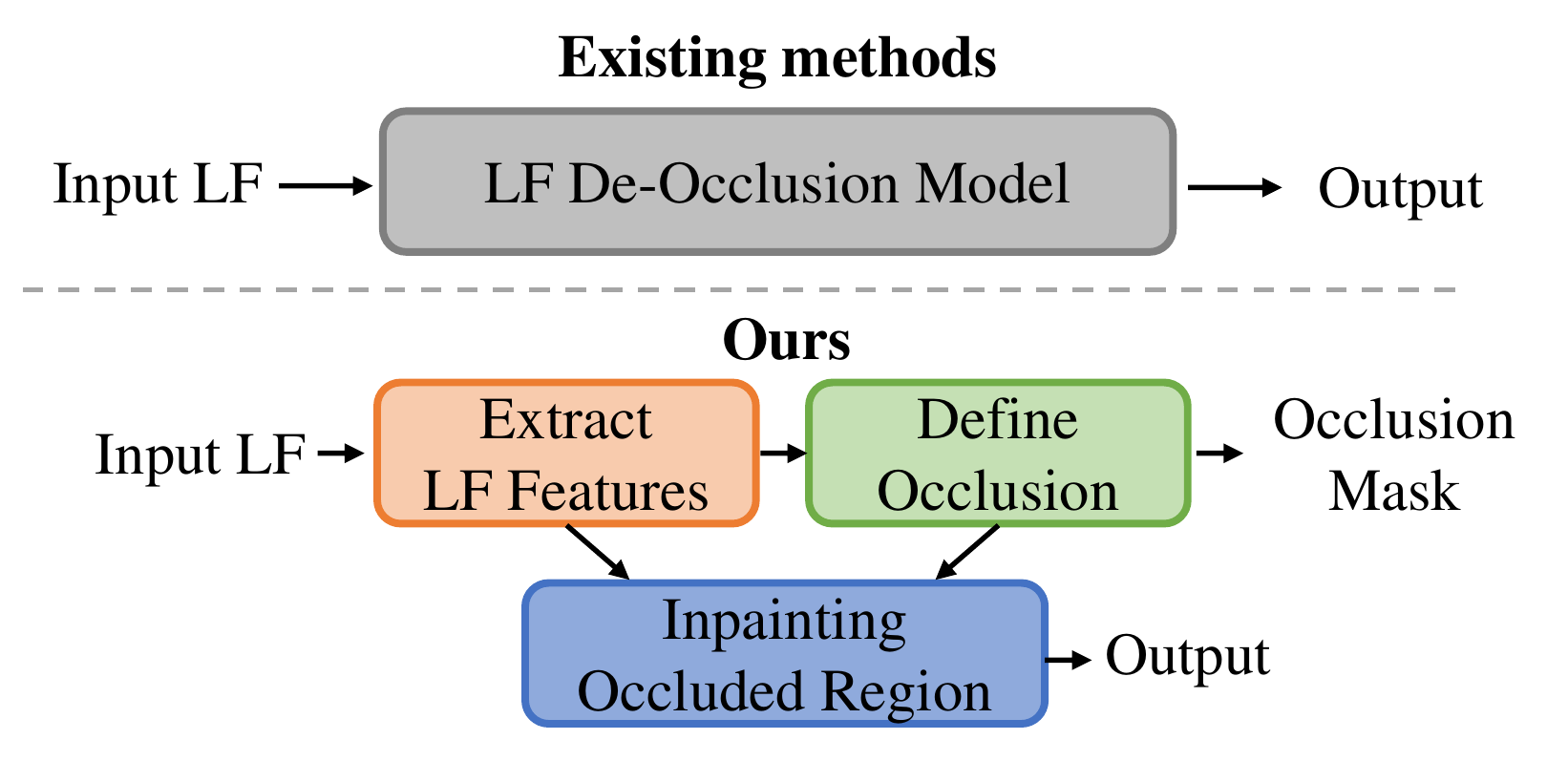}
\end{center}
    \caption{
    The framework of the existing LF-DeOcc methods (top) and the proposed method (bottom). 
    While the existing methods consist of a single black-box model, the proposed method is composed of three separate components, generating the occlusion mask as an intermediate representation of the framework which is useful for the analysis and other applications.
    Note that the proposed LF-DeOcc framework also works in an end-to-end manner.
    }
\label{fig:model_comparison}
\vspace{-2.5mm}
\end{figure}

In various computer vision tasks such as image classification \cite{resnet, vgg}, semantic segmentation \cite{unet, segnet,seg2}, and object detection \cite{yolo, yolo2, yolo3}, the performance becomes unstable and degrades by the foreground objects which occlude the region of interest.
To diminish such performance drop, the de-occlusion task aims to capture the foreground occlusion object in the image and fill the region with the backgrounds.

Recently, light fields (LFs) are utilized in the de-occlusion task (LF-DeOcc) \cite{deoccnet, ijcai2021-180}.
As LFs can capture the scene in various views with different angular information, they can offer guidance of the background object, the key goal of the de-occlusion task.
While the existing deep learning-based LF-DeOcc methods show reasonable performance, they mainly focus on the sparsely sampled (sparse) LFs, which are collected through a camera array and have large disparity ranges.
In other words, most of the occluded region can be seen in the other views of sparse LFs.
Contrarily, the densely sampled (dense) LFs are collected through a micro-lens-based \textit{portable} LF camera \cite{lytro_illum} which has narrow disparity ranges consisting of scarce information about the backgrounds.
In summary, while it is easy for the sparse LFs to obtain visible background information beyond the occlusion, it is hard for the dense LFs to obtain visible background information beyond the occlusion.
However, from a practical perspective, whereas it is easy for the dense LFs to obtain various scenes by being portable and affordable, it is relatively hard for the sparse LFs to collect various scenes.
The detailed statistics of the dataset and different characteristics of both LFs are provided in supplementary materials.

In this paper, different from the existing LF-DeOcc methods, we considered two scenarios to make a model work on both sparse and dense LFs. First, if the occluded region is visible in the other views, the visible information should be used to fill the region. Second, if not, the context information around the occluded region should be used to fill the region. 
Although both scenarios are presented in both LFs, the first scenario dominates in the sparse LFs whereas the second scenario dominates in the dense LFs.

Since each scenario requires quite different solutions, we divide the framework and define the separate roles and functions in the proposed framework (Fig.~\ref{fig:model_comparison}). For the first scenario, likewise DeOccNet \cite{deoccnet}, a component for extracting the LF features from the sub-aperture images (SAIs) is presented in the proposed framework. For the second scenario, a component for inpainting a single image is modified with an additional component to define the occlusion mask in the proposed framework.
Contrary to the existing methods which implicitly have those roles in their models, the proposed framework explicitly divides the roles and connects them. 
By explicitly dividing the roles, the proposed method not only shows better performance but also makes the development and analysis easier than the existing methods.

In addition, because the occlusion is explicitly represented in the proposed framework, it is flexible to define the occlusion, which helps not only prevent artifacts in the non-occluded regions but also remove occlusion in arbitrary depth planes while preserving the foreground objects of interest by manipulating the occlusion mask.

The contributions of the proposed LF-DeOcc framework are summarized as follows.
\vspace{-2.5mm}
\begin{itemize}
    \setlength{\itemsep}{-2pt}
    \setlength{\parsep}{-3pt}
    \item We propose a framework, ISTY, which works on both sparse and dense LFs, achieving \textit{state-of-the-art} performance in the majority of the settings.
    \item We modularized black-box framework into three separate components, making further development and analysis easier.
    \item Occlusion mask generator offers flexibility in defining the occlusion by explicitly giving the mask representation and enables additional applications.
\end{itemize}
\vspace{-2.5mm}
\section{Related work}

\subsection{Light Field De-Occlusion (LF-DeOcc)}
By the digital refocusing algorithm \cite{RenPhD}, in the refocused image, the occluded regions are blurry but partially visible. 
Thus, the digital refocusing algorithm has also been utilized to see through the occlusion \cite{Refocus}.
Vaish et al. \cite{Refocus} proposed a refocus method that re-parameterizes LF image by a specific value and average along with the angular dimension. In their method, even though foreground occlusion could be seen through, the images are highly blurred.

Recently, Wang et al. \cite{deoccnet} proposed a deep learning-based end-to-end LF-DeOcc model (DeOccNet).
DeOccNet reconstructs the occlusion-free center-view (CV) image with a deep encoder-decoder model and residual atrous spatial pyramid pooling (ResASPP) module from sparse LF images.
They also propose a mask embedding approach to generate a training dataset which synthesizes the occlusion LF image using the mask image and occlusion-free LF image allowing the fully supervised end-to-end LF-DeOcc learning.
However, DeOccNet generates blurry outputs and does not appropriately deal with occlusion with the large invisible region, making artifacts from occlusions.
Zhang et al. \cite{ijcai2021-180} proposed a filter to extract features from the shifted lenslet images to seek background information to reconstruct the occluded regions.
Although their method works well on the sparse LFs, the performance on the dense LF is not as good as that on the sparse LF because they strongly assume that background object is visible.
Furthermore, using a set of shifted-lenslet images requires large memory and long pre-processing time.

The recently proposed deep learning-based LF-DeOcc methods focused on the sparse LFs, filling the occluded regions with the visible background information from the other views.
Since it is more difficult to collect the sparse LF dataset than the dense LF dataset, it is reasonable to use the dense LFs, which can be easily collected, to train and apply a model with the advantage of a large number of data from practical perspectives.
Thus, different from existing LF-DeOcc methods, we propose a framework that works on both sparse and dense LF images.

\begin{figure*}[!ht]
\begin{center}
\includegraphics[width=0.9\textwidth]{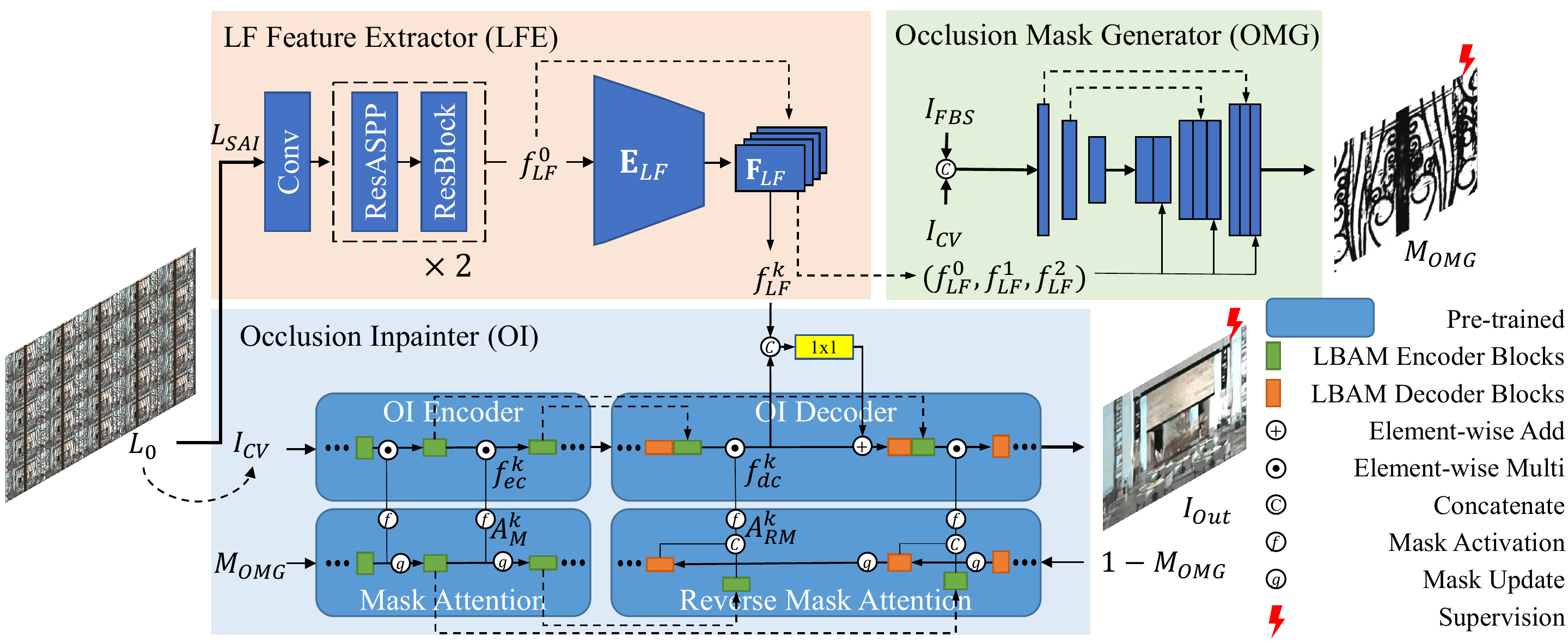}
\end{center}
   \caption{
    Illustration of the proposed model architecture. \raisebox{.5pt}{\textcircled{\raisebox{-.9pt} {$f$}}} and \raisebox{.5pt}{\textcircled{\raisebox{-.9pt} {$g$}}} denotes the mask activation and update function respectively, defined in LBAM \cite{LBAM}. The $I_{CV}$ and $M_{OMG}$ are supervised with the loss function defined in section \ref{sec:loss} using $I_{gt}$ and $M_{gt}$, respectively.
   }
\vspace{-2mm}
\label{fig:model_archi}
\end{figure*}

\subsection{Single Image Inpainting}\label{related:SII}

\textbf{Single image inpainting in an RGB image.} The goal of the single image inpainting is to recover the missing (masked) regions of a single image with realistic content. 
The rapid development of deep learning algorithms and the vast amount of single RGB image datasets \cite{psv, zhou2017places} makes it possible to fill the masked regions with a plausible structure without information beyond the mask \cite{Liu,RFR,LBAM}. 

Partial convolution (PConv) \cite{Liu} helps to encode the context features while avoiding the artifacts from invalid pixels of the masked region through the masked and re-normalized convolution.
Based on PConv, Li et al. \cite{RFR} introduced recurrent feature reasoning (RFR) to reconstruct the large continuous hole through recurrent inpainting the part of the image and average the generated feature group if they have no invalid pixels.
Xie et al. \cite{LBAM} use learnable bidirectional attention maps (LBAM) to replace the PConv.
They used attention not only in the encoder but also in the decoder so that the decoder can focus on filling the masked regions only.
In addition, unlike PConv, LBAM allows soft attention map and differentiable mask updates and activation function which gives more leeway for further improvements and stability to model in training.
However, directly applying the single image inpainting method on LF-DeOcc is not feasible as the occlusion mask can not be defined in the single RGB image.
In our framework, to utilize the inpainting method, we explicitly define the occlusion mask from the LFs and pass it as input of the inpainting method.

\textbf{Single image inpainting in LFs.} In the LF, the single image inpainting has been used to solve the LF completion which aims to fill the entire LF views with consistent information.
Rather than directly filling the 4-D manifold \cite{lf_inpaint_1,lf_inpaint_2,lf_inpaint_3}, Zhang et al.\cite{lf_inpaint_4} and Pendu et al.\cite{lf_inpaint_5} used single image inpainting to the CV image and propagated it to the remaining views.
However, since their main goal is to propagate the information of the inpainted CV image to the remaining views, they naively used the single image inpainting method for the CV image with a given inpainting mask.

\subsection{Foreground Background Seperation (FBS)}
The foreground-background separation (FBS) \cite{FBS-JSTSP, FBS-SFA, FBSOCC-SFA}, which represents whether each pixel is classified into the foreground or background, could be regarded as a sub-representation of the depth map \cite{LFUDM}.
Using the optical phenomenon \enquote{flipping} \cite{FBS-AO}, Lee and Park \cite{FBS-JSTSP} estimated depth map from LF by accumulating the binary maps, which separated the foreground and background based on the focused (or zero disparity) plane.
A single slice of FBS in form of the binary map is obtained by thresholding a score map, whose pixel values are bounded from $-1$ to $1$, with zero.
In this paper, we define the occlusion mask by refining the FBS in form of the score map, to avoid losing information.

\section{Method}

The proposed framework is composed of three components, each of which performs a different role, named LF feature extractor (LFE), occlusion mask generator (OMG), and occlusion inpainter (OI), respectively.
In the following subsections, we first describe our overall architecture.
Subsequently, we introduce each part of the model and the composition of our loss function.
Lastly, we introduce an additional application possible in the proposed framework.
The architecture of the proposed framework is shown in Fig.~\ref{fig:model_archi}.

\subsection{Overall Architecture} 
From the input LF $L_{0}\in \mathbf{R}^{U\times V\times X\times Y\times 3}$, we extract various LF information in LFE, define the occlusion mask in OMG, and reconstruct the occlusion-free CV image in OI, where $(U,V)$ and $(X,Y)$ denote the angular and spatial dimensions of the LF image respectively, and $3$ is color channel dimension.
In the LFE, using the $L_{0}$ in form of the SAIs concatenated along with the channel dimension, named $L_{SAI}\in \mathbf{R}^{(3 \times U\times V)\times X\times Y}$, a set of LF features $\mathbf F_{LF}=\{f_{LF}^{0}, \cdots , f_{LF}^{K}\}$ are extracted, where $K$ indicates the number of layers in LF encoder ($\mathbf{E}_{LF}$).
In the OMG, an FBS score map $I_{FBS}\in \mathbf{R}^{X\times Y\times 1}$ is directly obtained from $L_0$ without a deep learning-based method.
The occlusion mask $M_{OMG}\in \mathbf{R}^{X\times Y\times 1}$ is obtained by refining the $I_{FBS}$ with a CV image $I_{CV}$ and a set of LF features $\{f^0_{LF}, f^1_{LF}, f^2_{LF}\}$.
The OI reconstructs the occlusion-free CV image $I_{out}$ from $I_{CV}$ and $M_{OMG}$ in a single image inpainting manner.
We combine the LFE and OI by infusing the $\mathbf F_{LF}$ with OI decoder features $\mathbf F_{dc}$, to utilize the background information gathered from LFs during the inpainting step.
We train our model in a fully supervised manner using ground truth occlusion free CV image $I_{gt}$ and mask $M_{gt}$.

\subsection{LF Feature Extractor}
The main role of the LF feature extractor (LFE) is to find and extract rich information from $L_{SAI}$, including depth information, unoccluded background object information, and background context information.
To effectively handle large disparity objects scattered around the SAIs and extract context features of background information while avoiding large occlusion, LFE requires a large receptive field.

\textbf{LF feature initialization.} DeOccNet \cite{deoccnet} shows that the ResASPP module \cite{wang2019learning} is beneficial for large receptive field and helps to extract useful features required for LF-DeOcc task.
We use multi-layers of the ResASPP module and residual block (ResBlock) together for the large receptive field with a dense sampling rate, which is used in Wang et al.'s method \cite{lf-dfnet}.
We use $1 \times 1$ convolution followed by two ResASPP Block and ResBlock layers to initialize the LF feature ($f_{LF}^0$). 
In our model, ResASPP has four parallel dilated convolutions with dilation rates of 1, 2, 4, and 8, respectively, and our ResBlock consists of 3 convolution layers and two leakyReLU layers, alternately.

\textbf{LF feature encoding.} 
With the initialized feature $f_{LF}^0$, a set of LF features $f_{LF}^k (k>0)$ are extracted using a LF encoder ($\mathbf{E}_{LF}$).
$\mathbf{E}_{LF}$ consists of a $K$ encoder blocks, each of which consists of a convolution block followed by a self-attention module to give a more long-range dependency.
The convolution block uses 2D convolution with kernel size of 4, stride of 2, and padding size of 1, LeakyReLU and batch normalization.
Following the self-attention module which is defined in Zhang et al.~\cite{sagan}, we use a $1 \times 1$ convolution to generate the key, query, and value matrix from the output of the convolution block.
The output of the self-attention module is element-wisely added to the output of the convolution block with a learnable weight $\gamma$ which is initially set to 0.25.
For memory efficiency, only encoder layers that have direct skip connections to the OI use the self-attention module ($k>1$). 

\subsection{Occlusion Mask Generator}

The OMG generates occlusion mask $M_{OMG}$, from the $I_{FBS}$ and U-shaped refinement module. 
$I_{FBS}$ divides the foreground and background with respect to the zero disparity plane, and our 3-layer U-shaped network refines $I_{FBS}$ using the $I_{CV}$ as a guidance.
In the decoder part, we reuse the LF features extracted from LFE ($f_{LF}^k, k=(0,1,2)$) to efficiently take advantage of depth information encoded from LFE, by concatenating the $f_{LF}^k$ to the OMG decoder feature.
Finally, we generate an occlusion mask with a softmax layer, occlusion regions as 0, and non-occlusion regions as 1, ideally.
Note that contrary to the single image inpainting mask which is hard digit mask $\subset\{0,1\}$, our generated mask is soft continuous mask $\subset[0,1]$. 
Rather than thresholding the mask, which might cause a loss of information, we use the soft mask and use an appropriate inpainting method which can utilize the soft mask.

\subsection{Occlusion Inpainter}

\textbf{Inpainting Method.}
We use a U-shaped single image inpainting architecture for OI.
Contrary to PConv\cite{Liu} which only adopts hard 0-1 mask, mask attention used in LBAM~\cite{LBAM} can adopt soft mask due to the learnable attention map.
Thus, following the LBAM architecture \cite{LBAM}, the encoder feature of OI, $\tilde{f}_{ec}^k$, is re-normalized with mask attention $A^k_{M}$.
That is, $f_{ec}^k = \tilde{f}_{ec}^k \odot A_{M}^k$ where $\odot$ represents element-wise multiplication.
LBAM uses a mask attention map $A_M^k$ for encoder as well as reverse mask ($1-M_{OMG}$) attention map $A_{RM}^k$ for decoder.
The decoder feature of OI $\tilde{f}_{dc}^k$ is re-normalized with reverse mask attention, $f_{dc}^k = \tilde{f}_{dc}^k \odot A_{RM}^k$, which helps the decoder only focus on the masked region.

\textbf{Feature Fusion Method.}
One of the important functions of the OI is to fuse the decoder features of OI ($f_{dc}^k$) and $f_{LF}^k$ from LFE to reconstruct the occlusion-free CV image utilizing the visible background information from LFs.
We found that 1x1 convolution shows competitive or superior performance compared to other more complicated fusion methods.
Thus, for simplicity, we concatenate two features, $f_{LF}^k$ and $f_{dc}^k$, and infuse them with 1x1 convolution.
The fused features are element-wisely added to the $f_{dc}^k$ with a learnable parameter $\gamma$, which is initially set to 0.25.

\subsection{Loss Function}\label{sec:loss}
We follow the objective function used in LBAM \cite{LBAM} to guide the $I_{out}$ using the $I_{gt}$.
Our image reconstruction loss $\mathcal{L}_I$ consists of $\ell_1$ loss, perceptual loss and style loss, which are described in detail in LBAM \cite{LBAM},
\begin{equation}
    \mathcal{L}_I = \mathcal{L}_{\ell_{1}} + \lambda_1\mathcal{L}_{perc} + \lambda_2\mathcal{L}_{style}.
\end{equation}
Furthermore, with our modularized framework, the intermediate representation of the network, the occlusion mask $M_{OMG}$, can be directly guided using the $M_{gt}$.
Thus we add a mask generation loss $\mathcal{L}_M$ and use an MSE loss for it.
Finally, our entire objective function is defined as 

\vspace{-2.5mm}
\begin{equation}
    \mathcal{L} = \mathcal{L}_I + \mathcal{L}_M.
\end{equation}

\subsection{Arbitrary Depth Occlusion Removal} \label{sec:application}
By explicitly defining the occlusion mask, more flexible applications are available. 
In LF-DeOcc, the occlusion is defined as \textit{all objects placed in the foreground} based on the disparity plane $d$. 
However, in various real-world situations, the occlusion can be placed in some \textit{arbitrary} depth plane, and the object of interest may be placed in the foreground, which is undesired to be removed.
Different from previous researches, the proposed framework can only remove objects in an arbitrary depth plane while preserving the foreground object by simply manipulating the occlusion mask $M_{OMG}$, without additional finetuning.
Let the occlusion is placed between two disparity planes $d_1, d_2 (d_1<d_2)$.
The foreground occlusion mask for two disparity plane is denoted as $M_{OMG}^{d_1}, M_{OMG}^{d_2}$, respectively. 
Note that the occlusion mask $M_{OMG}^{d}$ defines all foreground occlusions between the disparity ranges $[d,\infty]$ as 0 and background as 1, ideally.
Then, the occlusion mask of objects placed between two disparity planes $[d_1,d_2]$ is defined as $M_{OMG}^{d_1,d_2} = 1-(M_{OMG}^{d_1}-M_{OMG}^{d_2})$.
With the $M_{OMG}^{d_1,d_2}$, the occlusion in an arbitrary depth plane can be removed, without affecting the foreground object.
In addition, unwanted occlusion removal can be prevented by the user definition.

\section{Experiments}\label{sec:exp}

\begin{figure*}[!ht]
\begin{center}
\includegraphics[width=\textwidth]{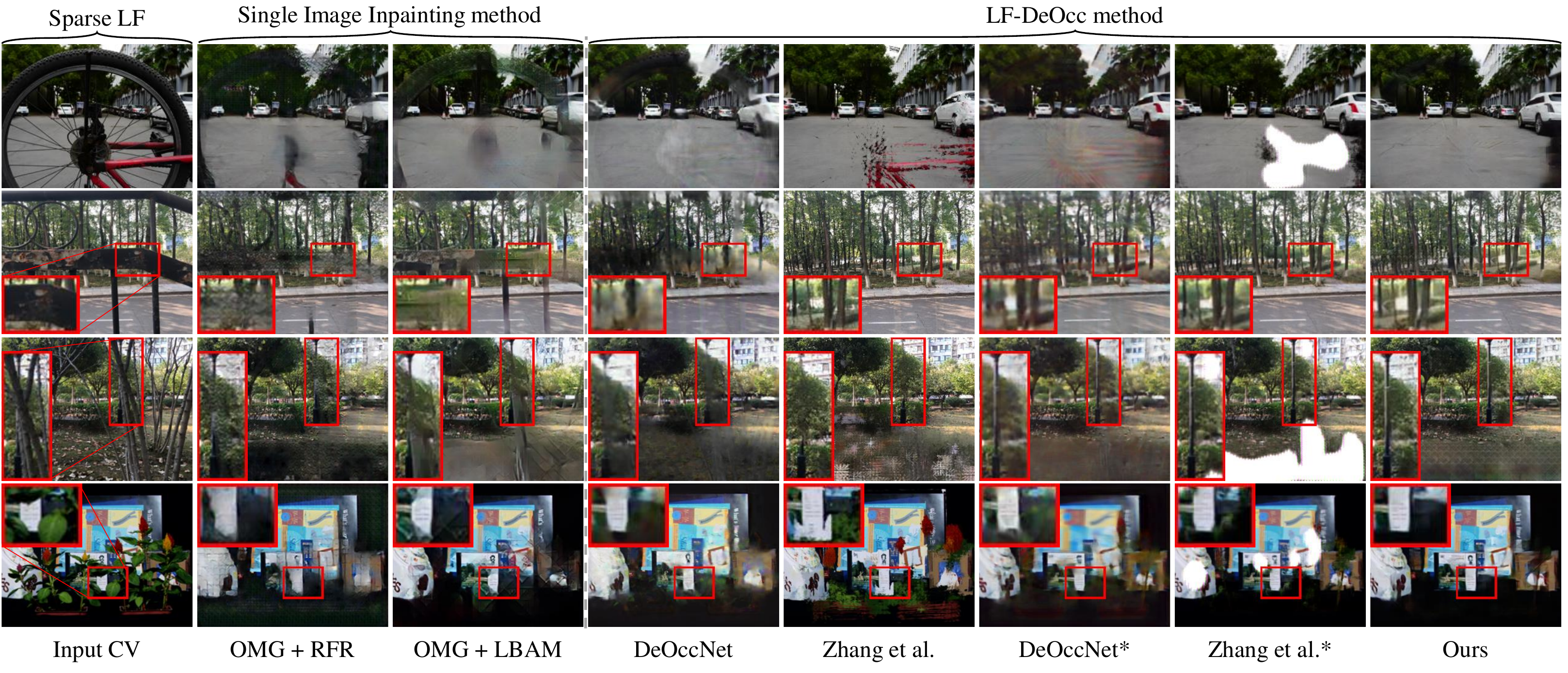}
\end{center}
   \caption{Qualitative comparisons on the \textit{sparse} LF dataset. Some parts of the outputs are magnified with red boxes for a detailed comparison.
   Ours could reconstruct sharper occlusion-free CV images from the scene utilizing occluded background information visible in other views. The scenes in the first last which is denoted as CD \cite{vaish2008new} are used for quantitative comparison.
   }
\label{fig:qual_sparse}
\vspace{-2.5mm}
\end{figure*}

\subsection{Experimental Setup} \label{exp_setup}

\textbf{Training dataset}
For the LF-DeOcc training, the requirement of the ground truth occlusion-free CV image is strong prior.
It has been researched that the mask embedding approach proposed by Wang et al. \cite{deoccnet} which synthesizes the occlusion LF image with occlusion-free LF image and occlusion image could solve this problem.
Even though it uses synthetic occlusion training data, the trained model could be well applied to various real-world occlusion scenes \cite{deoccnet, ijcai2021-180}.
The mask embedding approach randomly embeds 1-3 occlusion images in an LF image for a multi-disparity occlusion scenario.
To allow the model to effectively learn the inpainting scenario, that is, a low disparity occlusion scenario, we place more occlusions in the low disparity planes.
For the occlusion image, 21 thick and large real occlusion images are added to the original 80 mask images used in Wang et al.'s method \cite{deoccnet} to train the inpainting scenario caused by large occlusion.
A detailed description of the mask embedding approach that we applied and the disparity plane are provided in supplementary materials.

We embed the occlusion in the positive disparity planes.
Thus, our ground truth occlusion-free LFs should contain only negative disparity objects.
We choose 1418 LFs out of 2957 LFs from DUTLF-V2 \cite{DUTLF-V2} training dataset, which is a dense LF dataset captured by Lytro Illum camera~\cite{lytro_illum}.

\textbf{Test dataset}
To evaluate the performance quantitatively in sparse LFs, we use 4 synthetic sparse LF scenes (4-Syn) and 9 synthetic sparse LF scenes (9-Syn) for the quantitative comparison which is synthesized by Wang et al. \cite{deoccnet} and Zhang et al. \cite{ijcai2021-180}, respectively.
A real sparse LF image, Stanford CD scene \cite{vaish2008new} is also used for quantitative comparison as it has a ground truth.
To evaluate the performance quantitatively in dense LFs, we choose 615 LFs out of 1247 LFs from DUTLF-V2 \cite{DUTLF-V2} test dataset and collect another 33 real occlusion images.
Using the mask embedding approach with a disparity range of $[1, 4]$, single or double occlusions are embedded to evaluate multi-disparity occlusion scenario, which is denoted as \textit{Single Occ} and \textit{Double Occ} respectively.
For the qualitative comparison, various publicly available real-world sparse and dense occlusion LF scenes are used.
The sparse LF dataset is composed of LF scenes captured by Wang et al. \cite{deoccnet} and Stanford CD scene \cite{vaish2008new}.
The dense LF dataset is composed of EPFL-10 \cite{Rerabek:218363} and Stanford Lytro dataset \cite{stanford_dataset}, both captured by the Lytro Illum camera.

\textbf{Training Detail}
The angular and spatial resolution of LF images in DUTLF-V2\cite{DUTLF-V2} is $(U\times V\times X\times Y)=(9\times 9\times 600\times 400)$. 
For training and testing, we use central $5\times5$ images and the spatial resolution is resized to $300\times 200$. 
Randomly center-cropped and horizontally flipped images with a resolution of $(X\times Y)=(256\times 192)$ are used for our training with the mask embedding approach.
We randomly choose 1-3 masks with a random RGB shuffle and horizontally and vertically flipping for embedded occlusion masks in training time.
Our model is optimized by ADAM optimizer with ($\beta_1, \beta_2$)=(0.5,0.9), and a batch size, $\lambda_1$, and $\lambda_1$ are set to 16, 0.01, and 120.
The learning rate is initially set to 0.0005 and multiplied by 0.5 every 200 epochs. 
We train our model on 4 Nvidia TITAN X Pascal GPUs in Pytorch framework. 
The epoch is set to 500 and the training step is ended within 1 day. 

\subsection{Experimental Results}
We compare our model with the state-of-the-art LF-DeOcc methods, DeOccNet \cite{deoccnet} and Zhang et al.'s method \cite{ijcai2021-180}.
We train the DeOccNet with the same learning strategy and the dataset used by the original paper.
For Zhang et al.'s method, we used the pre-trained model provided by the author.
DeOccNet* and Zhang et al.* denotes each model trained on the same dataset and mask embedding as ours for a fair comparison in Dense LF dataset. 
We trained DeOccNet* from the scratch whereas Zhang et al.* is finetuned from pre-trained model provided by the author as the model trained from scratch does not converge.
We also compare our model with the single image inpainting methods, RFR \cite{RFR} and LBAM \cite{LBAM}, to investigate a information gathered from various views in LFs.
Since single image inpainting models can not define the foreground occlusions by itself, we additionally attach the OMG module to the single image inpainting model, where additional $(f_{LF}^0,f_{LF}^1,f_{LF}^2)$ are removed to prevent the information from LFs and $I_{FBS}$ are provided to define the occlusion.
Both RFR and LBAM pre-trained on the Paris Street View dataset \cite{psv} are finetuned on the same dataset we used.

\subsubsection{Qualitative Results} 
\begin{figure*}[!ht]
\begin{center}
\includegraphics[width=\textwidth]{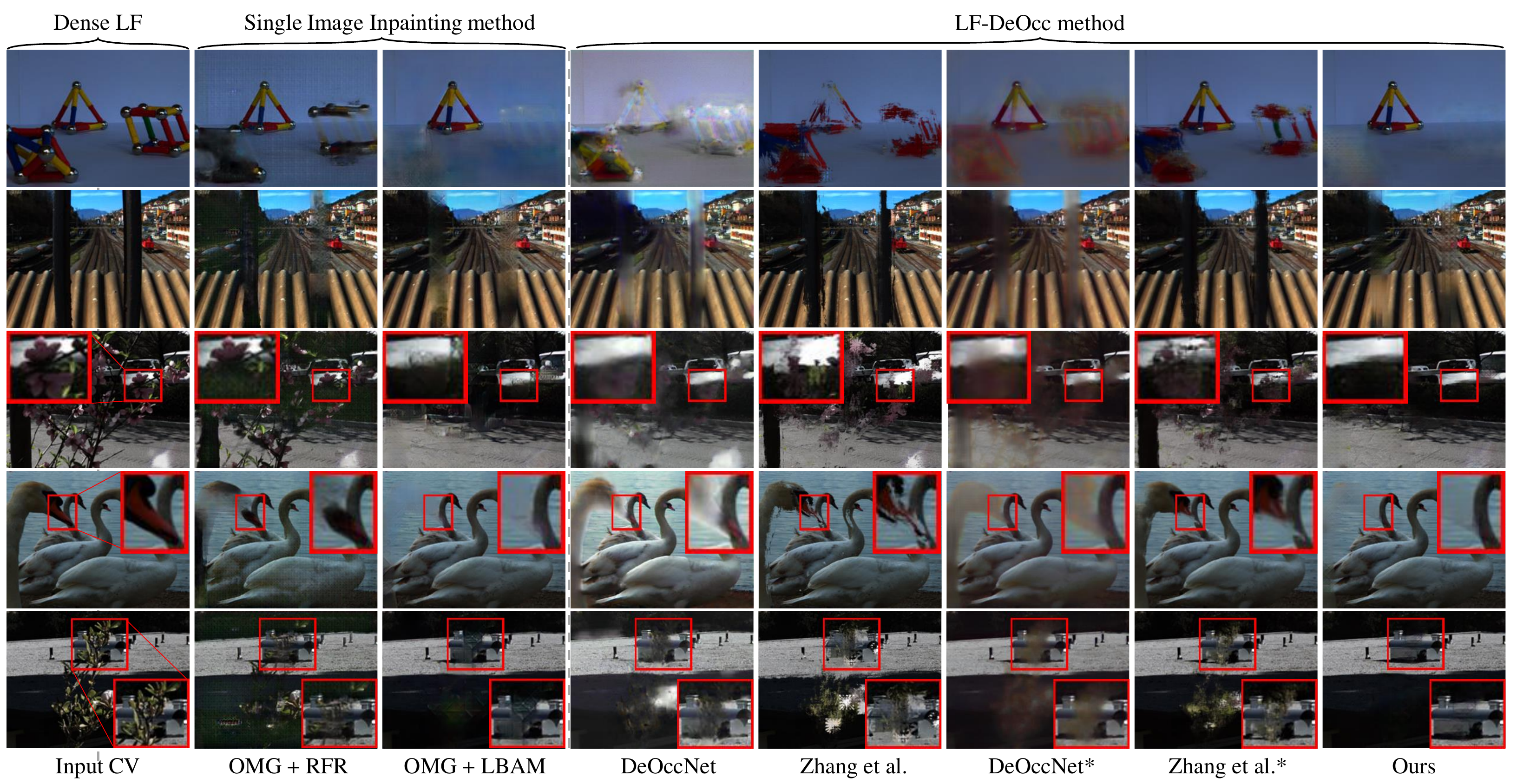}
\end{center}
   \caption{Qualitative comparisons on the \textit{dense} LF dataset. Some parts of the outputs are magnified with red boxes for a detailed comparison. Ours and OMG+LBAM\cite{LBAM} model can generate clear occlusion-free CV image.
   }
\label{fig:qual_dense}
\end{figure*}

\vspace{-2.5mm}
The qualitative comparisons of ours and other methods on real-world sparse and dense LFs are shown in Figs. \ref{fig:qual_sparse} and \ref{fig:qual_dense}.
The RFR \cite{RFR} does not reconstruct the scene in both LF datasets because PConv can not accept the soft mask generated by OMG.
With a learnable attention map, LBAM \cite{LBAM} outperforms in the dense LFs compared to RFR \cite{RFR} and existing LF-DeOcc methods. 
However, LBAM \cite{LBAM} can not accurately reconstruct the output in the sparse LFs since they entirely depend on the context information and can not accurately define the complicated occlusion without $\mathbf F_{LF}$.
We emphasize once again that RFR and LBAM do not work without OMG since the occlusion mask can not be defined in a single RGB image.
The DeOccNet \cite{deoccnet} shows blurry outputs around the occlusion in the sparse LFs and remains occlusion artifacts in both LF datasets.
DeOccNet* shows better reconstruction performance in both LF datasets, but still shows blurry outputs and remains occlusion artifacts.
Zhang et al.'s method \cite{ijcai2021-180} shows clear output if the single disparity occlusion has a large disparity (second row in Fig.~\ref{fig:qual_sparse}).
However, the artifacts from occlusions remain in the multi-disparity occlusions, especially in the dense LFs.
Even though they are trained on the same dataset that we used, Zhang et al.* \cite{ijcai2021-180} shows artifacts in the dense LFs.
Contrary to the single image inpainting models and other LF-DeOcc methods which only performs well in the dense and sparse LF datasets, respectively, our proposed framework generally shows better de-occlusion performance in both LF dataset.
Our model generates output with few occlusion artifacts compared to other LF-DeOcc models, reconstructing clear occlusion-free CV images.

\begin{table*}[!ht]
    \centering
    \caption{Quantitative comparison on the sparse and dense LF dataset using PSNR and SSIM (PSNR/SSIM). The higher the both metric, the better the quality of the reconstructed image. }
    {\footnotesize
    \begin{tabular}{c|c||c|c|c|c|c|c|c}
        LF Type & Name & RFR + OMG & LBAM + OMG & DeOccNet & {\scriptsize Zhang et al. \cite{ijcai2021-180}} & DeOccNet* & {\scriptsize Zhang et al.* \cite{ijcai2021-180}} & Ours \\ \hline\hline
        {\multirow{2}{*}{Sparse(syn)}} 
        & 4-Syn \cite{deoccnet}      & 19.89/0.668 & 21.11/0.677 & 25.04/0.807 & 26.37/\textbf{0.871} & 23.74/0.701 & 14.46/0.683 & \textbf{26.42}/0.836 \\
        & 9-Syn \cite{ijcai2021-180} & 20.69/0.672 & 23.04/0.725 & 21.07/0.791 & \textbf{27.97}/\textbf{0.901} & 23.70/0.715 & 22.00/0.758 & 27.04/0.849 \\ \hline
        {\multirow{1}{*}{Sparse(real)}} 
        & CD \cite{vaish2008new}     & 21.13/0.646 & 21.56/0.803 & 21.22/0.740 & 18.30/0.662 & 22.70/0.741 & 20.19/0.832 & \textbf{25.17}/\textbf{0.870}  \\ \hline
        {\multirow{2}{*}{Dense(syn)}}
        & \textit{Single Occ} & 26.28/0.867 & 27.92/0.899 & 24.84/0.863 & 21.98/0.815 & 28.67/0.914 & 23.15/0.900 & \textbf{32.44}/\textbf{0.947} \\
        & \textit{Double Occ} & 23.25/0.801 & 24.83/0.827 & 23.04/0.819 & 19.71/0.755 & 25.85/0.867 & 18.01/0.823 & \textbf{28.31}/\textbf{0.902}
    \end{tabular}
    }
    \label{quan_table}
\end{table*}

\vspace{-2.5mm}
\subsubsection{Quantitative Results}
We use peak signal-to-noise ratio (PSNR) and structural similarity index measure (SSIM) to quantitatively evaluate how precisely the model reconstructs the occlusion-free CV image, which is widely used metrics in LF-DeOcc and single image inpainting \cite{deoccnet, ijcai2021-180, LBAM, RFR}.

Table~\ref{quan_table} shows the summarized quantitative results.
The RFR \cite{RFR} and LBAM \cite{LBAM}, which is the single image inpainting method, shows better results in the dense LF images because the dense LF requires inpainting knowledge to reconstruct the occlusion-free scene. However, single image inpainting models generally shows lower performance in the sparse LFs because they can not utilize the background information from LFs.
DeOccNet \cite{deoccnet} and DeOccNet* shows reasonable results on both dataset, but generally shows insufficient performance.
As shown in Fig.~\ref{fig:qual_sparse}, Zhang et al. \cite{ijcai2021-180} shows notable performance, especially in the single disparity occlusion of the sparse LF.
Zhang et al.* \cite{ijcai2021-180} also does not generally shows better results.
Our proposed framework generally outperforms other LF-DeOcc and inpainting models in both sparse and dense LFs.

\subsection{Various Applications}
\subsubsection{Prevention of Unwanted Removal}
In the third row of the Fig.~\ref{fig:qual_sparse}, the unwanted regions may defined as an occlusion and removed, such as the ground, making serious artifacts.
Contrary to other LF-DeOcc models, by explicitly defining the occlusion mask, our model could manually prevent the unwanted removal with user guidance. Fig.\ref{fig:application} shows the original output $I_{out}$ and edited output $I_{out}^{edit}$ with the edited mask $M_{OMG}^{edit}$.
With explicit user guidance to the occlusion mask, the artifacts from the ground is removed in edited output.

\vspace{-2.5mm}
\subsubsection{Arbitrary Depth Occlusion Removal}
\begin{figure*}[!t]
\begin{center}
\includegraphics[width=0.88\textwidth]{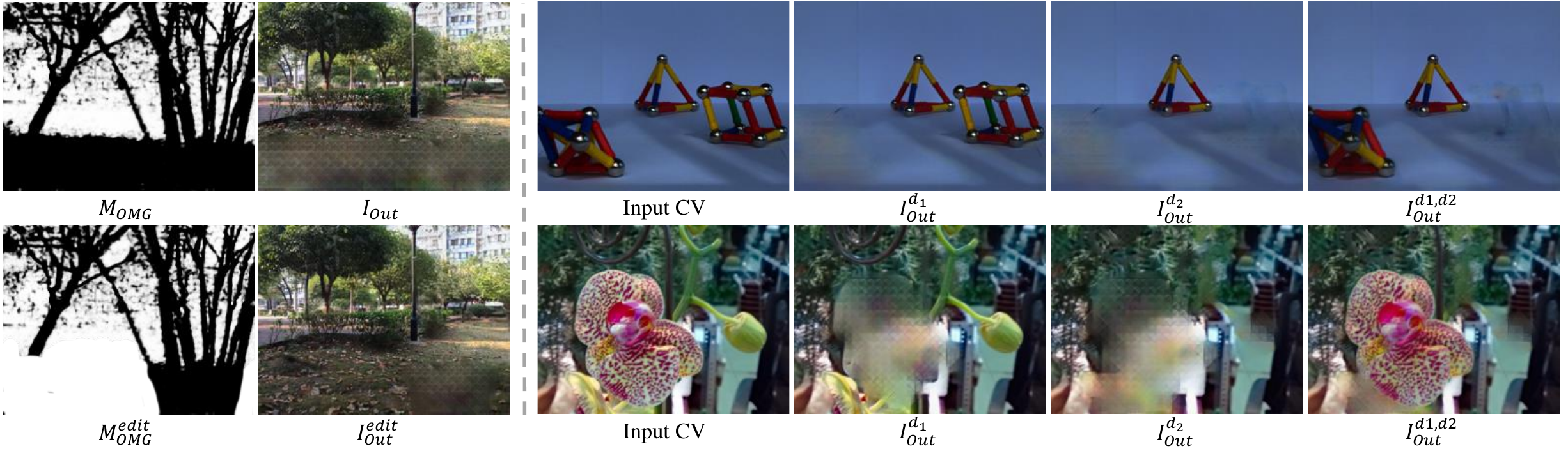}
\end{center}
  \caption{Illustration of the preventing unwanted removal (left) and arbitrary depth occlusion removal (right) in our model. Contrary to existing LF-DeOcc methods, our model could be applied to various de-occlusion tasks with mask manipulation.
  }
\label{fig:application}
\end{figure*}

Using the mask manipulation method described in section \ref{sec:application}, our model could be applied to the \textit{arbitrary depth occlusion removal}, which selectively removes the occlusion placed in arbitrary depth.
Fig. \ref{fig:application} shows the occlusion removal between two disparities $d_1$ and $d_2$.
The objects in the intermediate depth plane (parallelepiped-shaped magnet and flower bud) are removed while preserving the foreground objects (octahedron-shaped magnet and flower) without explicit guidance by the user.

\begin{table*}[!t]
    \centering
    \caption{Ablation studies of ours and its variants on the sparse and dense LF datasets using PSNR and SSIM (PSNR/SSIM). The higher the both metric, the better the quality of the reconstructed image.}
    {\small
    \begin{tabular}{c|c|c|c|c|c|c}
    LF Type& Name & \textit{DeOccNet-large} & \textit{DeOccNet-large + FBS} & \textit{SA Fusion} & \textit{M-F Fusion} & Ours \\ \hline\hline
    {\multirow{2}{*}{Sparse(syn)}}
    & 4-Syn\cite{deoccnet}               & 24.51/0.741 & 24.88/0.756 & 26.16/0.812 & 26.39/0.833 & \textbf{26.42}/\textbf{0.836} \\
    & 9-Syn\cite{ijcai2021-180}          & 24.24/0.755 & 25.08/0.774 & 26.79/0.847  & 26.89/\textbf{0.849}  & \textbf{27.04}/\textbf{0.849} \\ \hline
    Sparse(real) & CD \cite{vaish2008new} & 22.80/0.752 & 23.72/0.822 & 24.31/0.862 & 24.39/0.864  & \textbf{25.17}/\textbf{0.870} \\ \hline
    {\multirow{2}{*}{Dense(syn)}}
    & \textit{Single Occ}                & 29.39/0.927 & 31.20/0.939 & 32.18/0.945 & 32.06/0.944 & \textbf{32.44}/\textbf{0.947}\\
    & \textit{Double Occ}                & 26.44/0.881 & 27.76/0.896 & \textbf{28.35}/0.901 & 28.30/0.899 & 28.31/\textbf{0.902} 
    
    \end{tabular}
    }
    \label{tab:abla}
    \vspace{-2.5mm}
\end{table*}

\begin{table}[!t]
    \centering
    {\footnotesize
    \begin{tabular}{|c|c|c|c|c|} \hline
        Model       & LBAM \cite{LBAM} & Zhang et al.\cite{ijcai2021-180} & DeOccNet \cite{deoccnet} & Ours \\ \hline
        Params      & 69.3M & 2.7M & 39.0M & 80.6M \\ \hline
        $T_{inf}$ & 12ms & 3050ms & 10ms & 24ms \\ \hline
    \end{tabular}
    }
    \caption{The number of parameters and the average inference time $T_{inf}$ of each model. $T_{inf}$ is measured when calculating the LFs with spatial resolution of $256\times 192$ using a TITAN XP GPU.}
    \label{tab:param_table}
    \vspace{-2.5mm}
\end{table}

\subsection{Ablation Study}
The proposed framework is built upon the fact that LF-DeOcc requires various domain knowledge and by dividing the model into three specified components to deal with the sparse and dense LFs.
Since the three components are related closely, our framework does not properly work if one of the components is eliminated.
Thus, to verify the effectiveness of the separation, we design a \textit{DeOccNet-large}, which enlarges the DeOccNet \cite{deoccnet} in the channel dimension so that the number of the parameter is similar to ours (87.8M), but the model is not explicitly divided into specified components.
We further experiment \textit{DeOccNet-large + FBS}, in which $I_{FBS}$ is concatenated to the input $L_{SAI}$, to verify the effect of explicit guidance of the occlusion information on the performance.
Table \ref{tab:abla} shows the quantitative results of ablation studies.
With large parameters, \textit{DeOccNet-large} and \textit{DeOccNet-large + FBS} shows better performance in dense LFs, but still shows lower performance than ours.
Especially, even though they have a large number of parameters and explicit occlusion guidance, the performance improvement on the sparse LFs is insignificant and fails to generally deal with both sparse and dense LF datasets.
In addition, since our model combines single image inpainting methods and features of LFs, the feature fusion method affects the performance. 
We further test several attention-based fusion methods, self-attention based fusion \textit{(SA Fusion)} and mask-feature attention based fusion \textit{(M-F Fusion)}. The attention based fusion methods also outperform existing methods, but the performance improvement is marginal compared to the $1\times 1$ convolution we used, even though attention requires more parameters and computational powers. A detailed implementation of each fusion method is provided in supplementary materials.

\subsection{Limitations and Future Works}
The inpainting knowledge is necessary for LF-DeOcc in the dense LFs, requiring more parameters compared to only dealing with the sparse LF images.
Thus, our model is twice as large as DeOccNet \cite{deoccnet} (Table \ref{tab:param_table}).
However, ours shows reasonable inference time compared to other methods, which is appropriate for real-world applications (Table.~\ref{tab:param_table}).
For future work, more parameter-efficient LF-DeOcc methods are expected with efficient inpainting methods.
Furthermore, the inpainting knowledge of ours is sub-optimal because the inpainter is trained on a relatively small number of datasets compared to RGB dataset.
Pre-trained inpainter model does not alleviate this problem due to the catastrophic forgetting.
Some continual learning approaches \cite{ewc, gan_cont, piggyback} may be effective for this problem with a trade-off between the performance and memory, parameters, or training times.
Additionally, we expect combining our proposed framework with the LF completion methods~\cite{lf_inpaint_5} expands the LF-DeOcc from reconstructing single occlusion-free CV image to the entire occlusion-free LF image and gives new perspective to LF-DeOcc task.

\section*{Conclusion}
In this paper, we propose a deep learning-based LF-DeOcc framework, ISTY, which considers the various occlusion scenarios to work on both sparse and dense LF images.
By explicitly defining the occlusion mask and fusing background information from LF images into a single image inpainting model, the proposed framework can remove occlusions not only in the foreground but also in the arbitrary depth plane.
Various experimental results show that the proposed framework outperforms previous LF-DeOcc methods in both sparse and dense LF images, reconstructing clear occlusion-free images.
\textbf{Expected Societal Impact.} Since the proposed framework can remove the objects in arbitrary depth plane, without affecting other objects, it could be abused to conceal the crime scene or hide crucial clues.

\textbf{Acknowledgements.}
This work was supported by Institute of Information \& communications Technology Planning \& Evaluation (IITP) grant funded by the Korea government(MSIT) (No.2022-0-00866, Development of cyber-physical manufacturing base technology that supports high-fidelity and distributed simulation for large-scalability) and National Research Foundation of Korea (NRF) grant funded by the Korea government(MSIT) (No. 2022R1A2C201270611).

{\small
\bibliographystyle{ieee_fullname}
\bibliography{egbib}
}

\newpage
\clearpage
\appendix

\begin{figure*}[!t]
\begin{center}
\includegraphics[width=\textwidth]{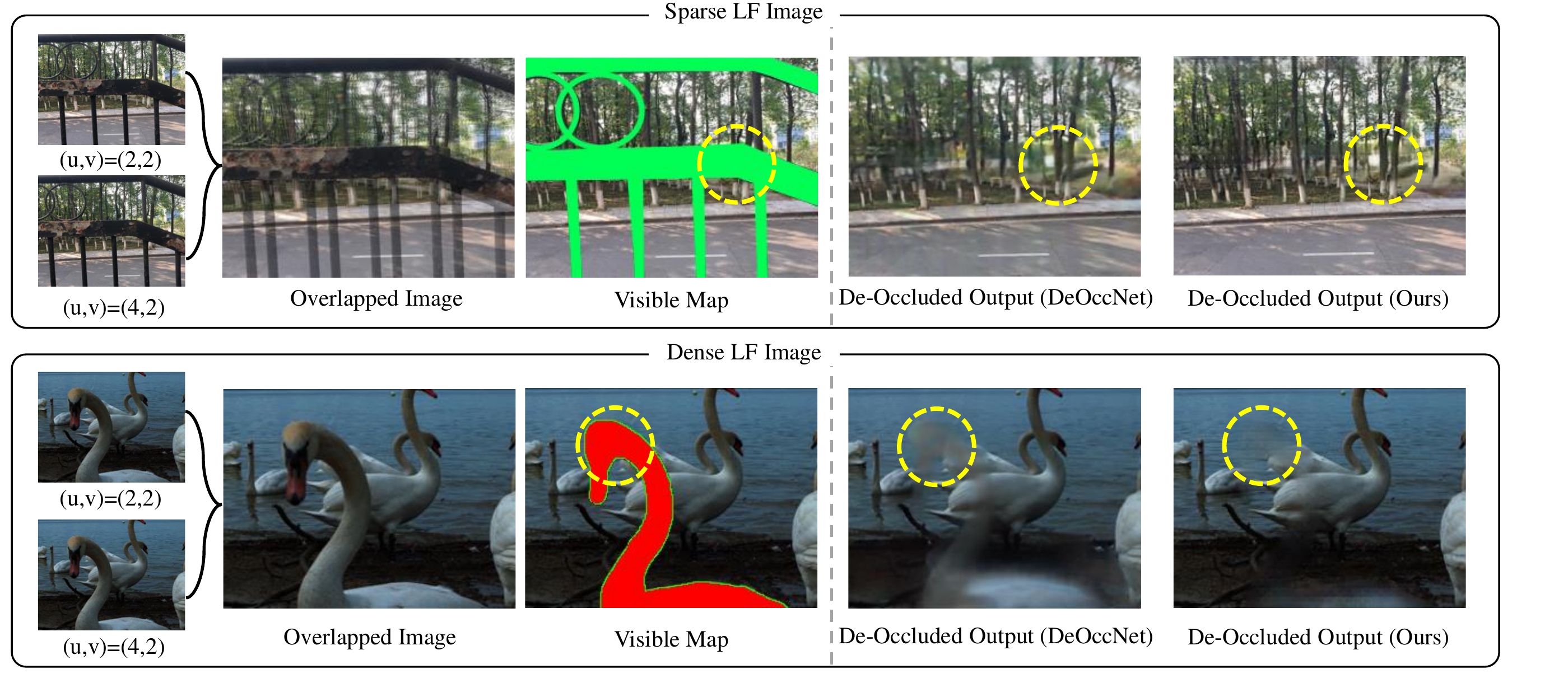}
\end{center}
   \caption{Illustration of the different characteristics of sparse LF image (top) and the dense LF image (bottom).}
\label{fig:dense_sparse}
\end{figure*}

\section{Comparison between Sparsely and Densely Sampled Light Fields}

Fig. \ref{fig:dense_sparse} shows the different characteristics between sparsely sampled (sparse) and densely sampled (dense) LF images.
We draw the overlapped images between center-view (CV) images and the rightmost images of LF. 
Both LF images have the angular resolution $(U, V) = (5,5)$, where $U$ and $V$ represent the horizontal and vertical angular resolutions. 
The corresponding angular coordinate $(u,v)$ is $(2,2)$ and $(4,2)$, respectively. 
It is clear that the occlusion in the sparse LF image shows larger disparities than occlusion in the dense LF image. 
In the visible map, with respect to the CV images $(u,v)=(2,2)$, while the green label denotes the occluded regions visible in other views, the red label signifies the occluded regions invisible in other views. We labeled the visible maps by ourselves. 
In the yellow dotted circle, the dense LF image has more invisible regions even both LF images have a similar occlusion size.
On the right of gray dotted lines, there are de-occlusion outputs of DeOccNet \cite{deoccnet} and ours, both trained on the dense LF dataset.
The proposed model could reconstruct clear occlusion-free CV image with reduced occlusion artifacts in both sparse and dense LF images.

\begin{table}[!b]
\begin{center}
\begin{tabular}{|c|c|c|}
\hline
Category & Dataset & \# of Scenes \\
\hline\hline
{\multirow{2}*{Sparse LF}} & Stanford & 30 \\
                        & DeOccNet Train \cite{deoccnet} & 60 \\ \hline
{\multirow{3}*{Dense LF}}  & DUTLF \cite{DUTLF} & 1462 \\
                        & DUTLF-V2 \cite{DUTLF-V2} & 4204 \\
                        & LFSD \cite{lfsd} & 100 \\
\hline
\end{tabular}
\end{center}
\caption{The number of training and test scenes of publicly available real-world LF datasets. Dense LF datasets usually have larger number of scenes compared to sparse LF datasets.}
\label{tab:num_dset}
\end{table}

The table \ref{tab:num_dset} shows the comparison of the number of scenes in the sparse and dense LF datasets.
Clearly, the dense LF datasets have the larger number of scenes than the sparse LF datasets because it is easier to collect the dense LF scenes using the portable LF camera than the sparse LF scenes. 
Thus, it is reasonable to train a model using the dense LF datasets to make the model learn various features in scenes.

\section{Experiments on Various Fusion Methods}

\begin{figure*}[!t]
\begin{center}
\includegraphics[width=\textwidth]{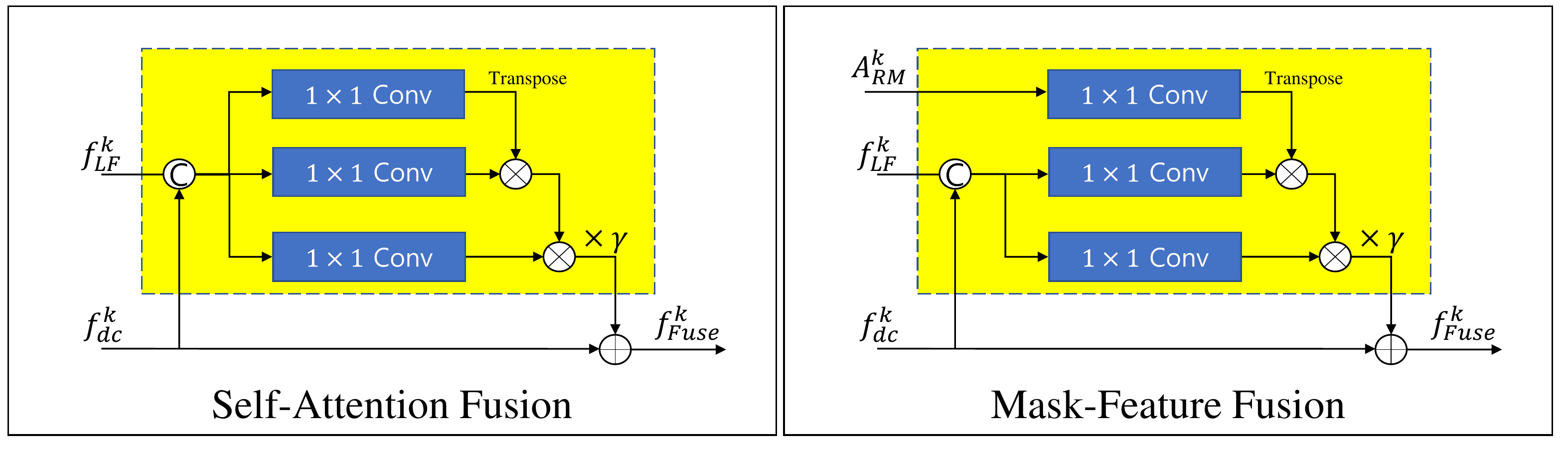}
\end{center}
   \caption{Illustration of the fusion methods used in ablation study, self-attention based fusion and mask-feature based fusion.}
\label{fig:fusion}
\end{figure*}

The proposed framework combines the LF features ($\mathbf F_{LF}$) to the decoder features of occlusion inpainter ($\mathbf F_{dc}$) to reconstruct the occlusion-free CV image through the background information from LFs as well as context information.
Since the $\mathbf F_{LF}$ includes not only useful background information but also occlusion information, which may cause artifacts, a careful fusion method is required.
We mainly focus on the attention based feature fusion methods to filter out occlusion artifact from $\mathbf F_{LF}$ and only background information is combined to fused features $f_{Fuse}$.
The self-attention, which is used in LF feature extractor (LFE) of our proposed framework, is experimented as self-attention fusion (\textit{SA fusion}).
The output features of $k^{th}$ layer of encoder in LFE ($f_{LF}^k$) is concatenated to the $k^{th}$ layer of decoder in occlusion inpainter ($f_{dc}^k$).
The self-attention output is calculated by the concatenated features and residually added to the $f_{dc}^k$ with learnable parameter $\gamma$, which is initially set to 0.25.
Our occlusion inpainter (OI) has reverse mask attention in the decoder ($A_{RM}$) which has feature-level information about the occlusion.
Inspired by an encoder-decoder attention used in the Transformer~\cite{transformer}, in which the decoder refers the encoder feature to generate a sentence, we design a mask-feature attention which refers the occlusion mask information to fuse the features. (\textit{M-F fusion}).
Based on the self-attention module, we replace the input of key convolution as $A_{RM}^{k}$, the $k^{th}$ layer of reverse mask attention.
Figure~\ref{fig:fusion} shows the detailed architecture of each fusion methods.
Even though attention based fusion methods requires more parameters and computational power, simple 1x1 convolution shows similar performance. Since 1x1 convolution is more efficient and could be easily applied to other architectures, we adopt 1x1 convolution as a fusion method for proposed framework.

\section{Mask Embedding Method}
\subsection{Light Field Reparameterization}
Light field (LF) reparameterization \cite{LFreparam,FBS-AO} can be expressed as 
\begin{align}
        L_d(x,y,u,v) = L_0(x+ud,y+vd,u,v),
        \label{eq:reparam}
\end{align}
where $L_d$ and $L_0$ signify the reparameterized LF and input LF images, respectively. 
By controlling the disparity plane $d$, the zero-disparity plane (focused plane) can be moved.
In the mask embedding of our training step, a single occlusion mask is copied to each view image, then the copied multiple occlusion masks are reparameterized on an arbitrary disparity plane $d$. 
By doing so, a mask can have the disparity information in the 4-D LF manifold.

\subsection{Settings for Reparameterization}
\begin{figure*}[!t]
    \small
    \centering
    \begin{tabular}{cccccc}
        \rotatebox[origin=c]{90}{$d=0.00$} &
        \includegraphics[width=.18\textwidth,align=c]{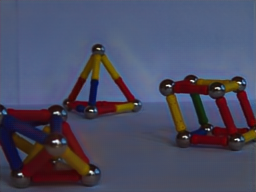} \hspace{-3.7mm} & 
        \includegraphics[width=.18\textwidth,align=c]{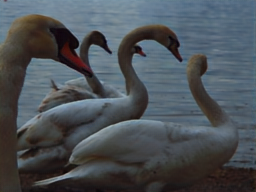} \hspace{-3.7mm} & 
        \includegraphics[width=.18\textwidth,align=c]{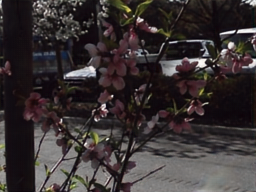} \hspace{-3.7mm} & 
        \includegraphics[width=.18\textwidth,align=c]{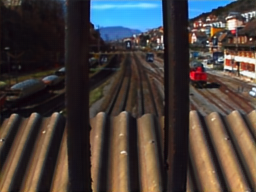} \hspace{-3.7mm} & 
        \includegraphics[width=.18\textwidth,align=c]{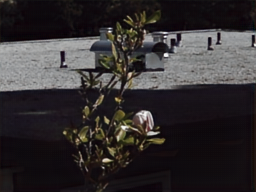} \hspace{-3.7mm} \\ \\
        \rotatebox[origin=c]{90}{$d=0.25$} &
        \includegraphics[width=.18\textwidth,align=c]{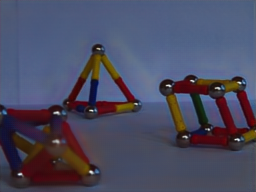} \hspace{-3.7mm} & 
        \includegraphics[width=.18\textwidth,align=c]{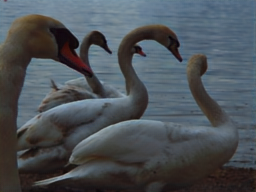} \hspace{-3.7mm} & 
        \includegraphics[width=.18\textwidth,align=c]{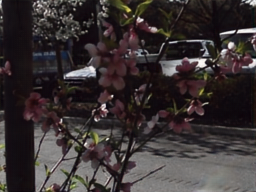} \hspace{-3.7mm} & 
        \includegraphics[width=.18\textwidth,align=c]{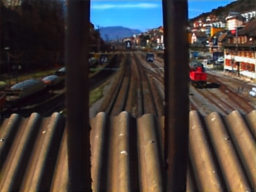} \hspace{-3.7mm} & 
        \includegraphics[width=.18\textwidth,align=c]{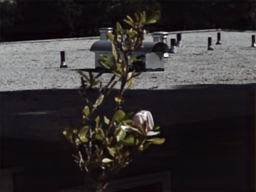} \hspace{-3.7mm} \\ \\
        \rotatebox[origin=c]{90}{$d=0.50$} &
        \includegraphics[width=.18\textwidth,align=c]{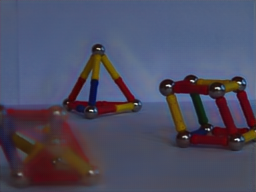} \hspace{-3.7mm} & 
        \includegraphics[width=.18\textwidth,align=c]{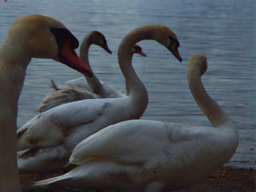} \hspace{-3.7mm} & 
        \includegraphics[width=.18\textwidth,align=c]{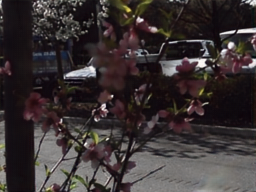} \hspace{-3.7mm} & 
        \includegraphics[width=.18\textwidth,align=c]{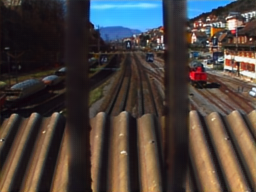} \hspace{-3.7mm} & 
        \includegraphics[width=.18\textwidth,align=c]{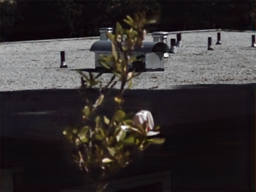} \hspace{-3.7mm} \\ \\
        \rotatebox[origin=c]{90}{$d=0.75$} &
        \includegraphics[width=.18\textwidth,align=c]{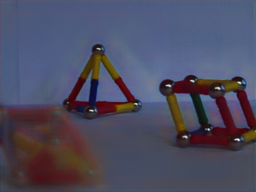} \hspace{-3.7mm} & 
        \includegraphics[width=.18\textwidth,align=c]{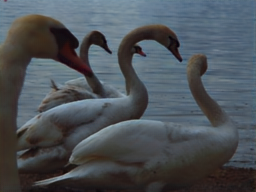} \hspace{-3.7mm} & 
        \includegraphics[width=.18\textwidth,align=c]{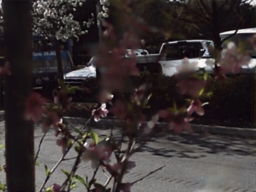} \hspace{-3.7mm} & 
        \includegraphics[width=.18\textwidth,align=c]{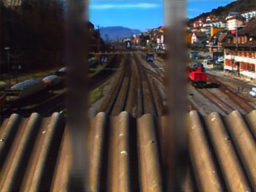} \hspace{-3.7mm} & 
        \includegraphics[width=.18\textwidth,align=c]{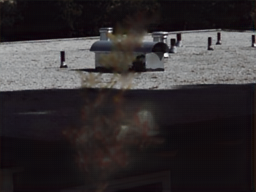} \hspace{-3.7mm} \\ \\
        \rotatebox[origin=c]{90}{$d=1.00$} &
        \includegraphics[width=.18\textwidth,align=c]{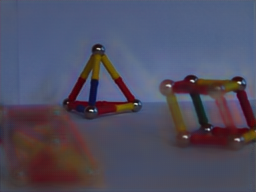} \hspace{-3.7mm} & 
        \includegraphics[width=.18\textwidth,align=c]{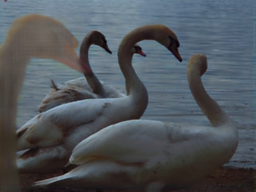} \hspace{-3.7mm} & 
        \includegraphics[width=.18\textwidth,align=c]{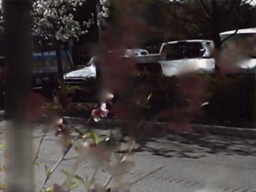} \hspace{-3.7mm} & 
        \includegraphics[width=.18\textwidth,align=c]{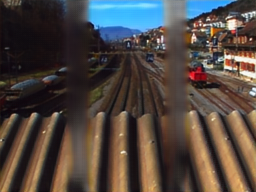} \hspace{-3.7mm} & 
        \includegraphics[width=.18\textwidth,align=c]{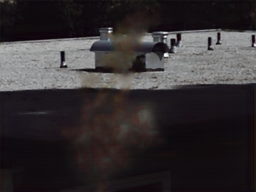} \hspace{-3.7mm} \\ \\
        \rotatebox[origin=c]{90}{$d=1.25$} &
        \includegraphics[width=.18\textwidth,align=c]{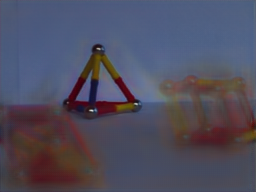} \hspace{-3.7mm} & 
        \includegraphics[width=.18\textwidth,align=c]{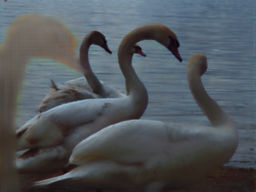} \hspace{-3.7mm} & 
        \includegraphics[width=.18\textwidth,align=c]{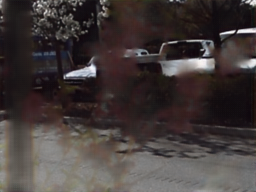} \hspace{-3.7mm} & 
        \includegraphics[width=.18\textwidth,align=c]{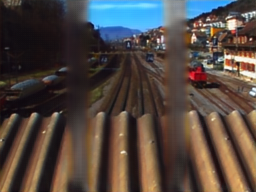} \hspace{-3.7mm} & 
        \includegraphics[width=.18\textwidth,align=c]{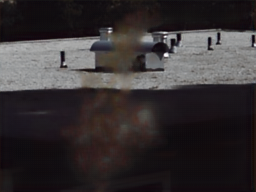} \hspace{-3.7mm} \\ \\ 
        \rotatebox[origin=c]{90}{$d=1.50$} &
        \includegraphics[width=.18\textwidth,align=c]{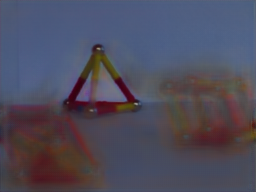} \hspace{-3.7mm} & 
        \includegraphics[width=.18\textwidth,align=c]{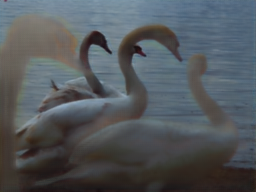} \hspace{-3.7mm} & 
        \includegraphics[width=.18\textwidth,align=c]{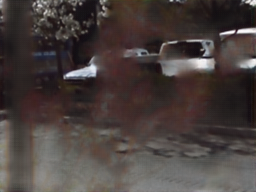} \hspace{-3.7mm} & 
        \includegraphics[width=.18\textwidth,align=c]{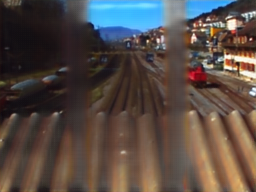} \hspace{-3.7mm} & 
        \includegraphics[width=.18\textwidth,align=c]{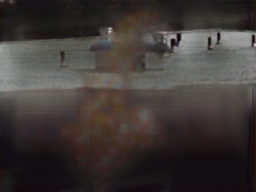} \hspace{-3.7mm}
    \end{tabular}
    \caption{Different outputs from the DeOccNet*. Each row shows the output LF-DeOcc image depending on the input LF image reparameterized by the parameter $d$. }
    \label{fig:reparam_deoccnet}
\end{figure*}

\begin{figure*}[!t]
    \small
    \centering
    \begin{tabular}{cccccc}
        \rotatebox[origin=c]{90}{$d=0.00$} &
        \includegraphics[width=.18\textwidth,align=c]{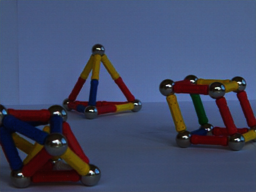} \hspace{-3.7mm} & 
        \includegraphics[width=.18\textwidth,align=c]{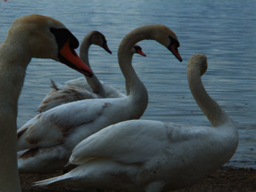} \hspace{-3.7mm} & 
        \includegraphics[width=.18\textwidth,align=c]{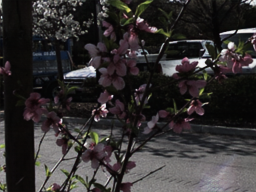} \hspace{-3.7mm} & 
        \includegraphics[width=.18\textwidth,align=c]{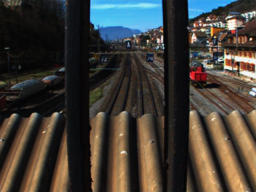} \hspace{-3.7mm} & 
        \includegraphics[width=.18\textwidth,align=c]{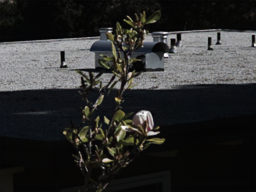} \hspace{-3.7mm} \\ \\
        \rotatebox[origin=c]{90}{$d=0.25$} &
        \includegraphics[width=.18\textwidth,align=c]{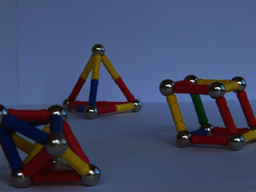} \hspace{-3.7mm} & 
        \includegraphics[width=.18\textwidth,align=c]{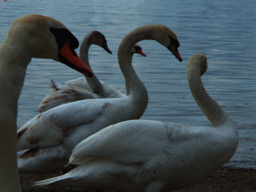} \hspace{-3.7mm} & 
        \includegraphics[width=.18\textwidth,align=c]{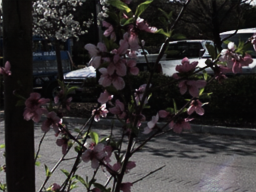} \hspace{-3.7mm} & 
        \includegraphics[width=.18\textwidth,align=c]{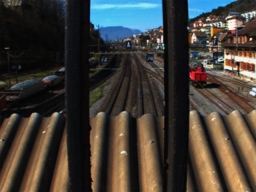} \hspace{-3.7mm} & 
        \includegraphics[width=.18\textwidth,align=c]{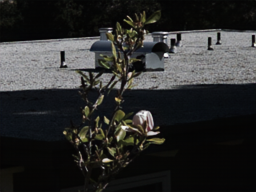} \hspace{-3.7mm} \\ \\
        \rotatebox[origin=c]{90}{$d=0.50$} &
        \includegraphics[width=.18\textwidth,align=c]{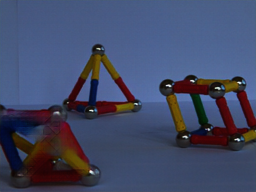} \hspace{-3.7mm} & 
        \includegraphics[width=.18\textwidth,align=c]{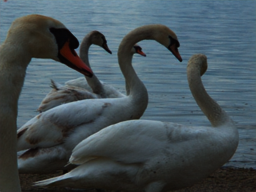} \hspace{-3.7mm} & 
        \includegraphics[width=.18\textwidth,align=c]{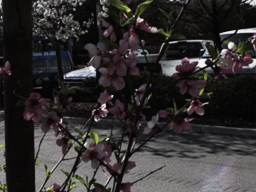} \hspace{-3.7mm} & 
        \includegraphics[width=.18\textwidth,align=c]{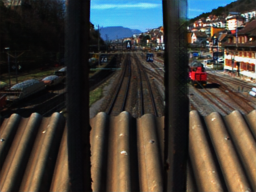} \hspace{-3.7mm} & 
        \includegraphics[width=.18\textwidth,align=c]{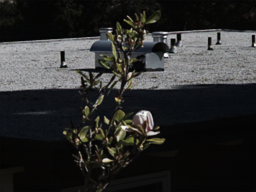} \hspace{-3.7mm} \\ \\
        \rotatebox[origin=c]{90}{$d=0.75$} &
        \includegraphics[width=.18\textwidth,align=c]{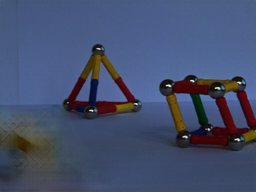} \hspace{-3.7mm} & 
        \includegraphics[width=.18\textwidth,align=c]{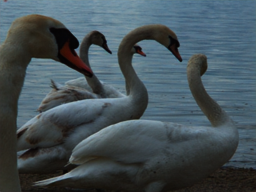} \hspace{-3.7mm} & 
        \includegraphics[width=.18\textwidth,align=c]{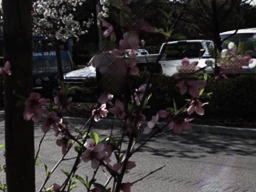} \hspace{-3.7mm} & 
        \includegraphics[width=.18\textwidth,align=c]{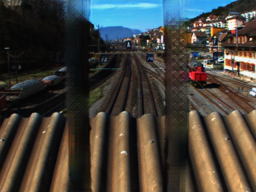} \hspace{-3.7mm} & 
        \includegraphics[width=.18\textwidth,align=c]{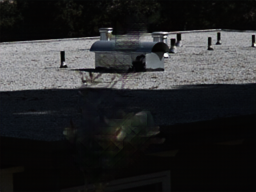} \hspace{-3.7mm} \\ \\
        \rotatebox[origin=c]{90}{$d=1.00$} &
        \includegraphics[width=.18\textwidth,align=c]{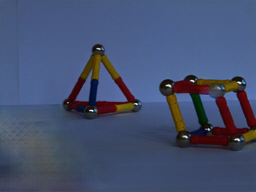} \hspace{-3.7mm} & 
        \includegraphics[width=.18\textwidth,align=c]{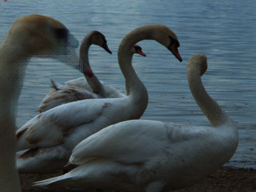} \hspace{-3.7mm} & 
        \includegraphics[width=.18\textwidth,align=c]{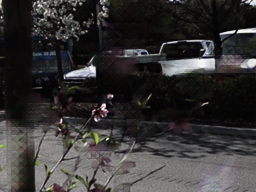} \hspace{-3.7mm} & 
        \includegraphics[width=.18\textwidth,align=c]{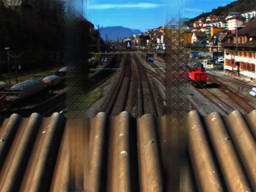} \hspace{-3.7mm} & 
        \includegraphics[width=.18\textwidth,align=c]{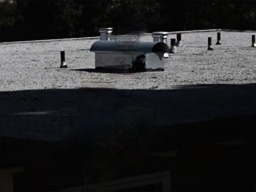} \hspace{-3.7mm} \\ \\
        \rotatebox[origin=c]{90}{$d=1.25$} &
        \includegraphics[width=.18\textwidth,align=c]{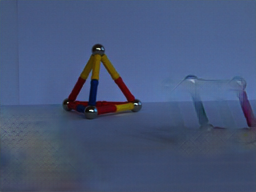} \hspace{-3.7mm} & 
        \includegraphics[width=.18\textwidth,align=c]{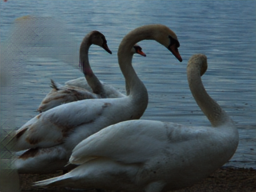} \hspace{-3.7mm} & 
        \includegraphics[width=.18\textwidth,align=c]{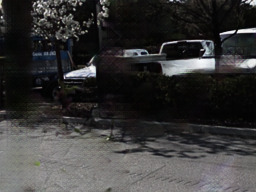} \hspace{-3.7mm} & 
        \includegraphics[width=.18\textwidth,align=c]{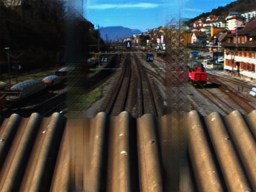} \hspace{-3.7mm} & 
        \includegraphics[width=.18\textwidth,align=c]{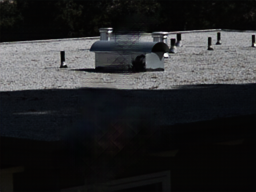} \hspace{-3.7mm} \\ \\
        \rotatebox[origin=c]{90}{$d=1.50$} &
        \includegraphics[width=.18\textwidth,align=c]{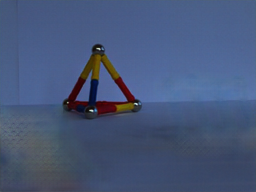} \hspace{-3.7mm} & 
        \includegraphics[width=.18\textwidth,align=c]{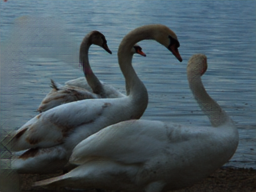} \hspace{-3.7mm} & 
        \includegraphics[width=.18\textwidth,align=c]{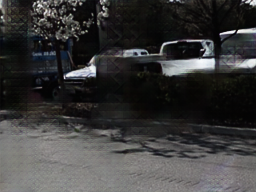} \hspace{-3.7mm} & 
        \includegraphics[width=.18\textwidth,align=c]{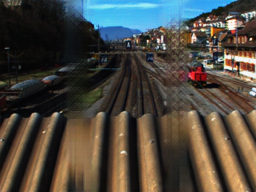} \hspace{-3.7mm} & 
        \includegraphics[width=.18\textwidth,align=c]{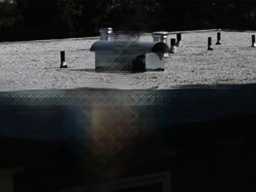} \hspace{-3.7mm}
    \end{tabular}
    \caption{Different outputs from the proposed framework. Each row shows the output LF-DeOcc image depending on the input LF image reparameterized by the parameter $d$.}
    \label{fig:reparam_ours}
\end{figure*}

In the qualitative results on the dense LFs in the main manuscript, we control the zero-disparity plane of input LF images to make the foreground occlusions have positive disparity, by reparameterizing the LF scenes from EPFL-10 \cite{epfl_dataset} and Stanford Lytro dataset \cite{stanford_dataset}.
Figs.~\ref{fig:reparam_ours} and \ref{fig:reparam_deoccnet} show the output de-occluded images generated by the DeOccNet* and the proposed framework from various disparity planes $d$ in Eq.~\ref{eq:reparam}, where DeOccNet* denotes the DeOccNet~\cite{deoccnet} trained on the same training dataset with ours.
If $d$ is too small, foreground objects are not considered as occlusion and if $d$ is too large, background objects are also considered as occlusion.
With the proper LF reparametrization, the foreground occlusion is properly removed with the proposed framework.

\subsection{Detailed Implementation of Mask Embedding} \label{supp:mask_embed}
In this subsection, we describe the detailed mask embedding approach used in this paper.
Although the existing methods \cite{deoccnet} generated the mask embedded scenes before training as pre-pocessing, we generate various scenes using a set of occlusion mask templates in training time for the data augmentation.

Original mask embedding randomly embed the 1-3 masks to deal with multi disparity occlusion scenario, and they randomly select the disparity plane $d$ from $[0, D]$, $[D, 2D]$, and $[2D, 3D]$ for the first, second and third occlusion.
Different from the existing methods, we embed more occlusion in low disparity plane in order that the model could deal with dense LFs.
At the same time, the model should be also trained on a large disparity occlusion scenarios to work on the sparse LFs.
Therefore, we randomly select 1-3 occlusions with disparity plane from $[0,1]$, $[1,4]$, and $[4,9]$, respectively, not uniformly selecting the disparity plane.

\section{Evaluation Details}
\begin{figure*}[!ht]
    \centering \includegraphics[]{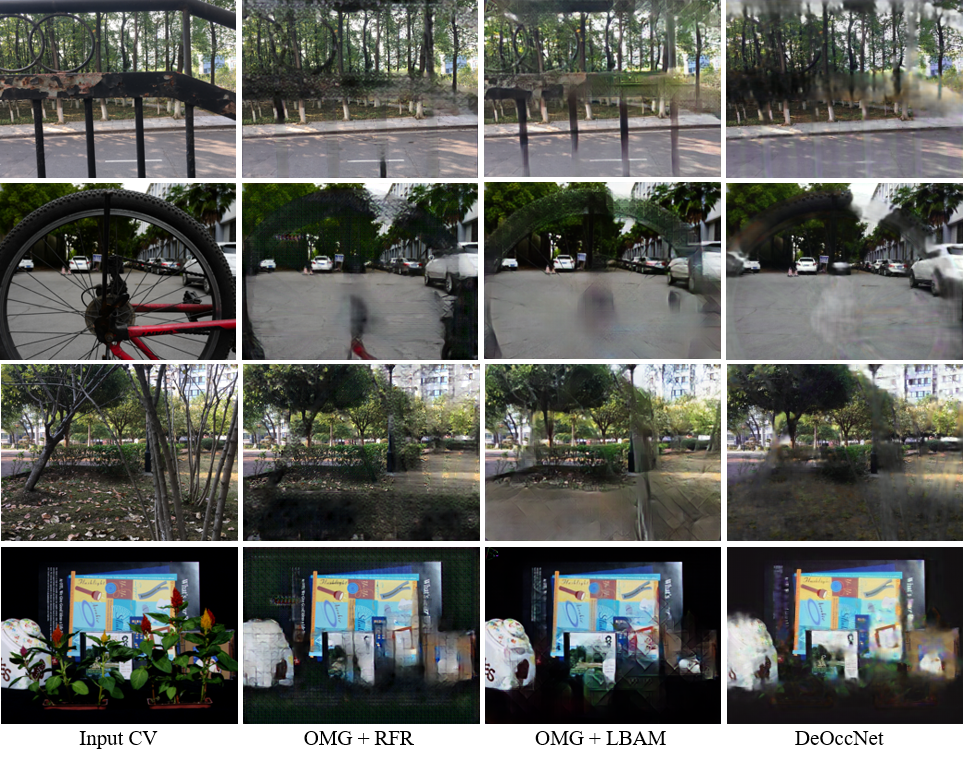}
    \caption{The enlarged outputs used in the qualitative results of main manuscript.}
    \label{fig:t1}
\end{figure*}

\begin{figure*}[!ht]
    \centering \includegraphics[]{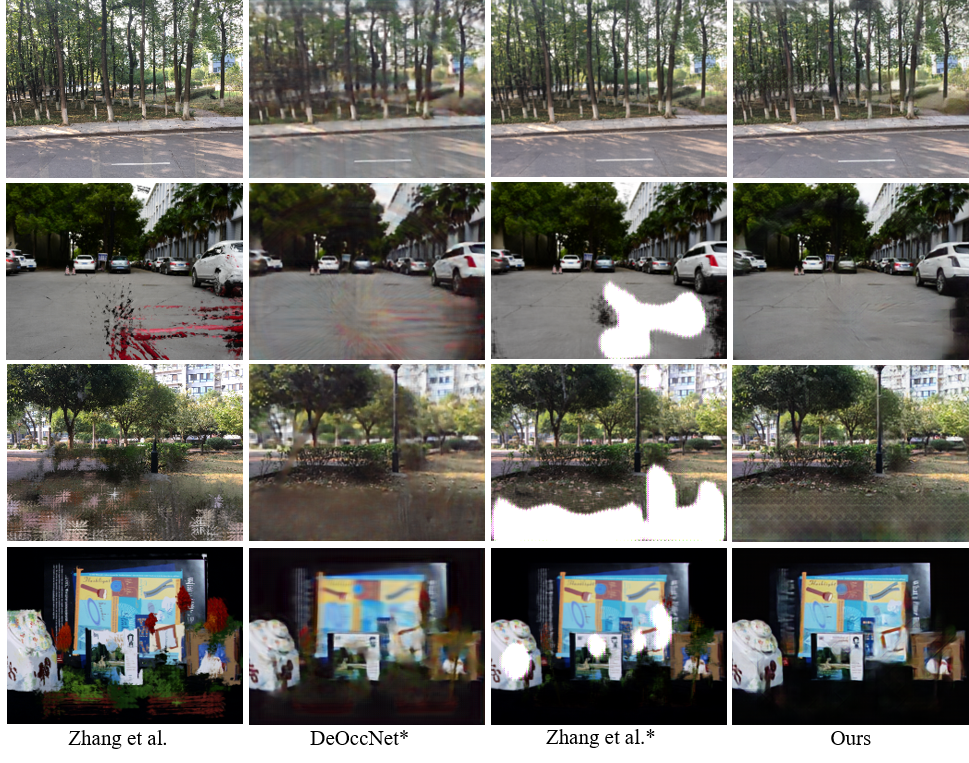}
    \caption{The enlarged outputs used in the qualitative results of main manuscript.}
    \label{fig:t2}
\end{figure*}

\begin{figure*}[!ht]
    \centering \includegraphics[]{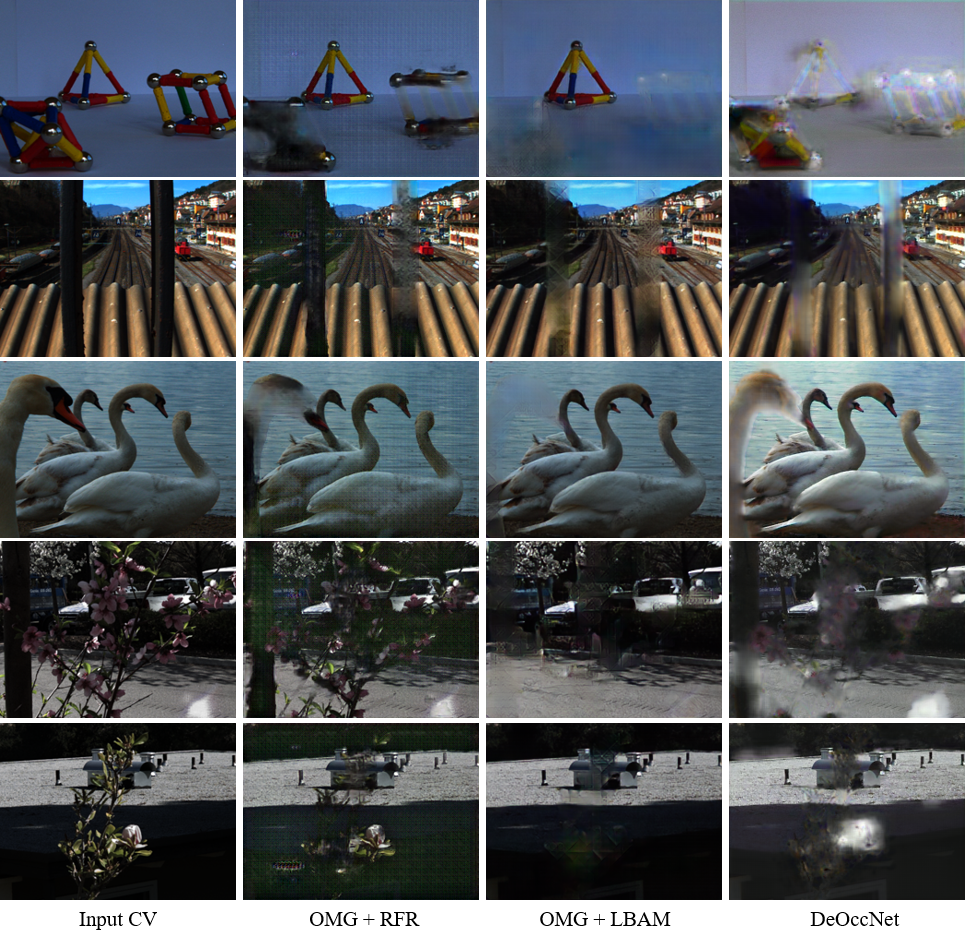}
    \caption{The enlarged outputs used in the qualitative results of main manuscript.}
    \label{fig:t3}
\end{figure*}

\begin{figure*}[!ht]
    \centering \includegraphics[]{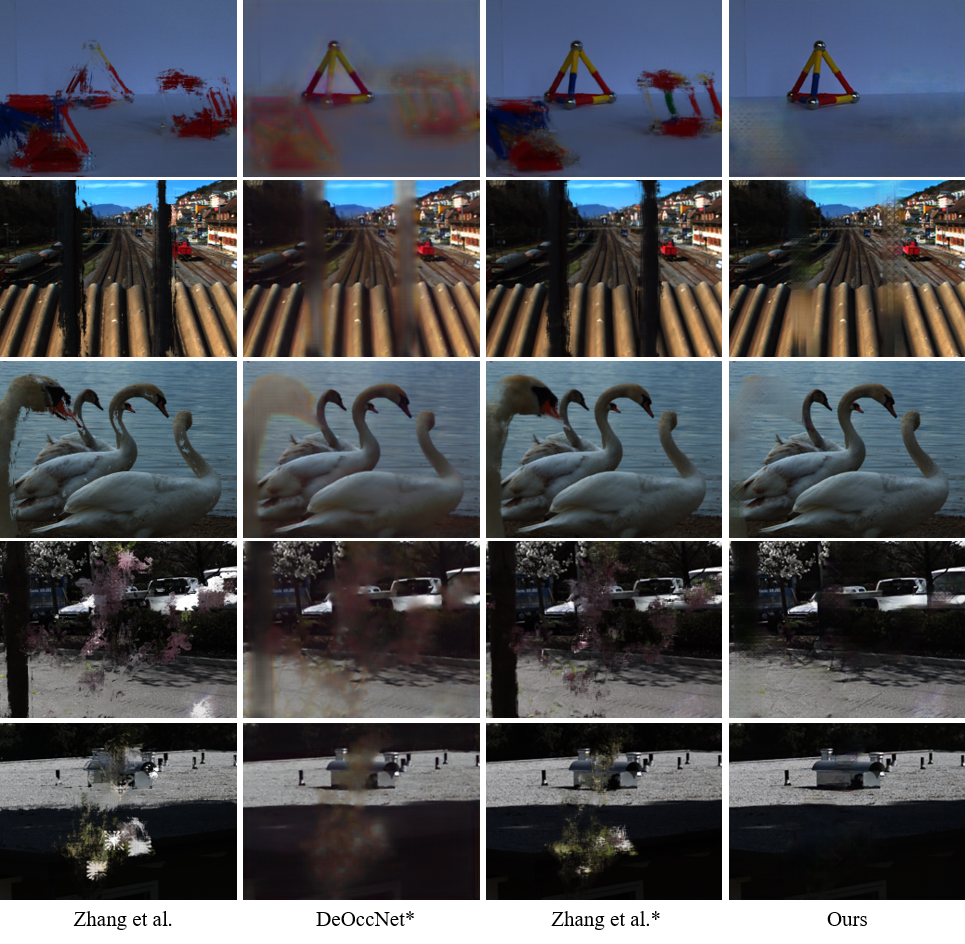}
    \caption{The enlarged outputs used in the qualitative results of main manuscript.}
    \label{fig:t4}
\end{figure*}

\textbf{Qualitative results.}
In this part, we provide the enlarged input, output images used in qualitative results of the paper for a detailed comparison (Figs.~\ref{fig:t1}, \ref{fig:t2}, \ref{fig:t3}, and \ref{fig:t4}).

\begin{figure*}[!t]
\begin{center}
\includegraphics[width=\textwidth]{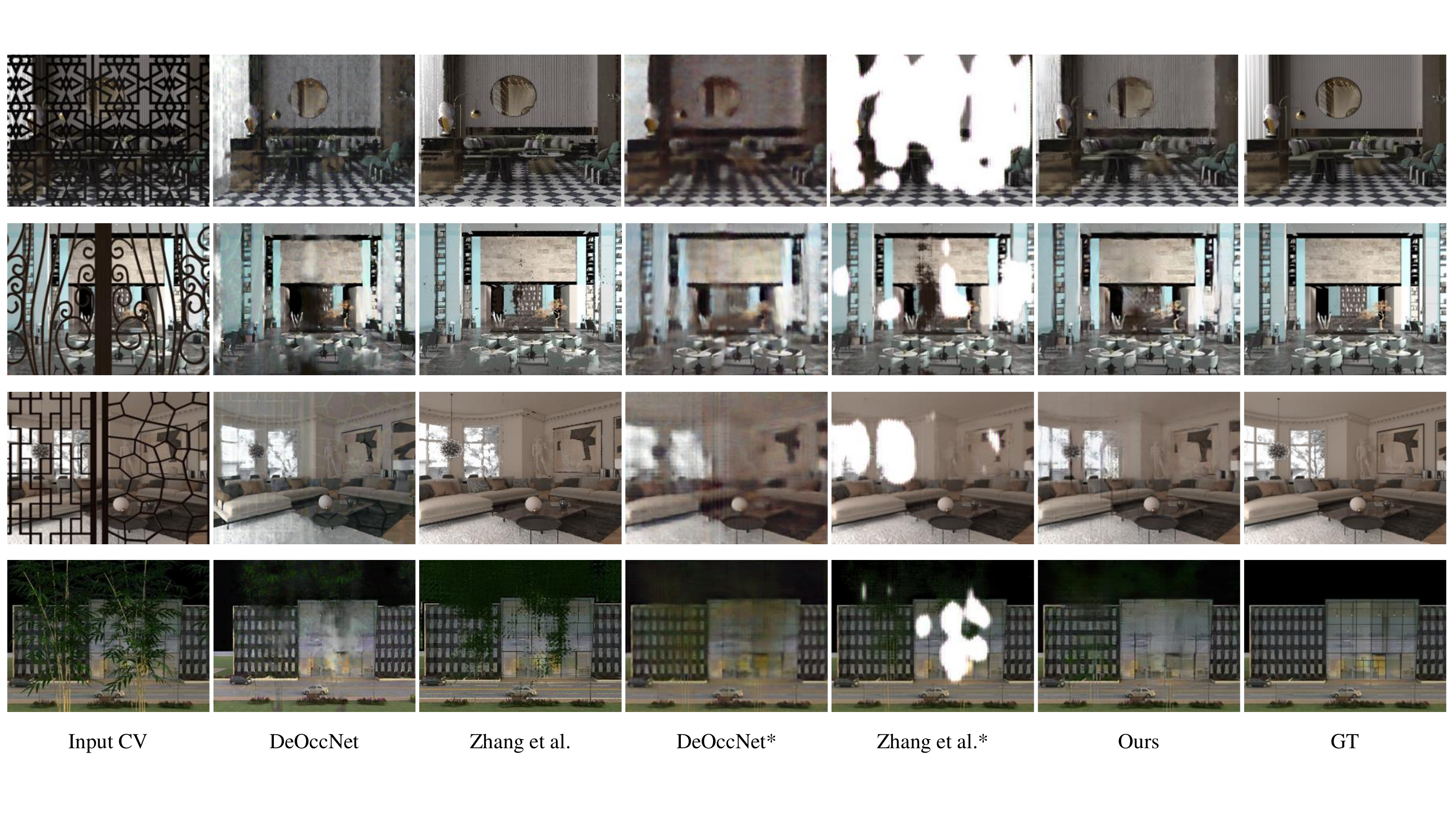}
\end{center}
   \caption{De-occlusion outputs on 4-syn dataset\cite{deoccnet} using various LF-DeOcc methods, which is used for quantitative results. }
\label{fig:4syn_outputs}
\end{figure*}

\begin{figure*}[!h]
    \small
    \centering
    \begin{tabular}{ccccccc}
        \includegraphics[width=.134\textwidth]{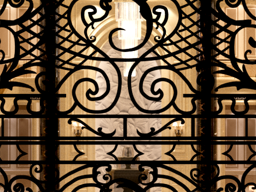} \hspace{-4mm} & 
        \includegraphics[width=.134\textwidth]{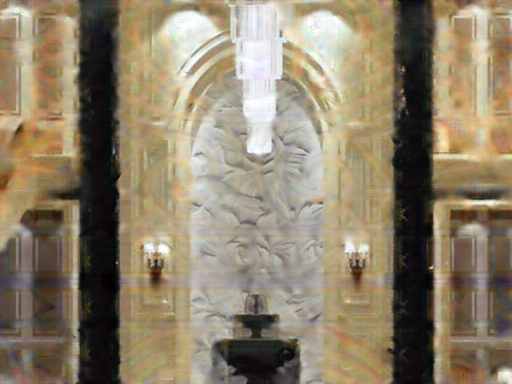} \hspace{-4mm} & 
        \includegraphics[width=.134\textwidth]{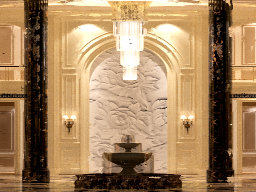} \hspace{-4mm} & 
        \includegraphics[width=.134\textwidth]{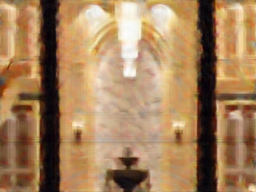} \hspace{-4mm} & 
        \includegraphics[width=.134\textwidth]{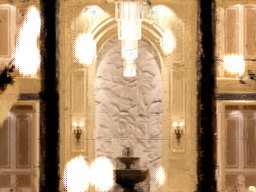} \hspace{-4mm} & 
        \includegraphics[width=.134\textwidth]{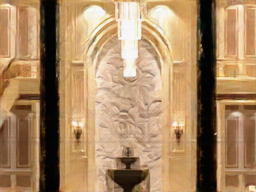} \hspace{-4mm} &
        \includegraphics[width=.134\textwidth]{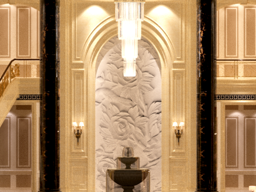} \hspace{-4mm} \\
        \includegraphics[width=.134\textwidth]{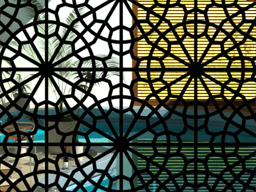} \hspace{-4mm} & 
        \includegraphics[width=.134\textwidth]{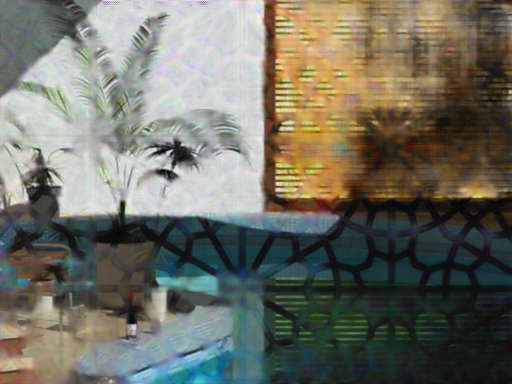} \hspace{-4mm} & 
        \includegraphics[width=.134\textwidth]{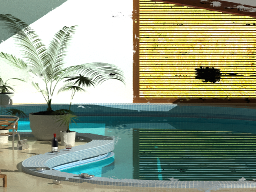} \hspace{-4mm} & 
        \includegraphics[width=.134\textwidth]{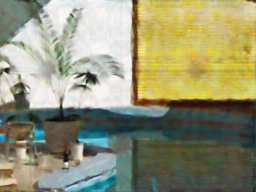} \hspace{-4mm} & 
        \includegraphics[width=.134\textwidth]{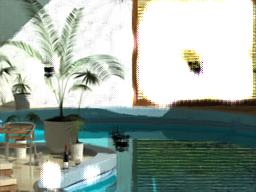} \hspace{-4mm} & 
        \includegraphics[width=.134\textwidth]{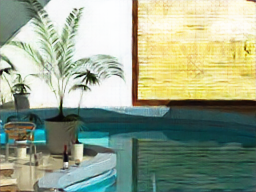} \hspace{-4mm} &
        \includegraphics[width=.134\textwidth]{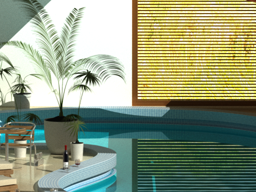} \hspace{-4mm} \\
        \includegraphics[width=.134\textwidth]{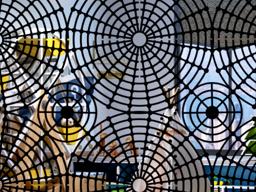} \hspace{-4mm} & 
        \includegraphics[width=.134\textwidth]{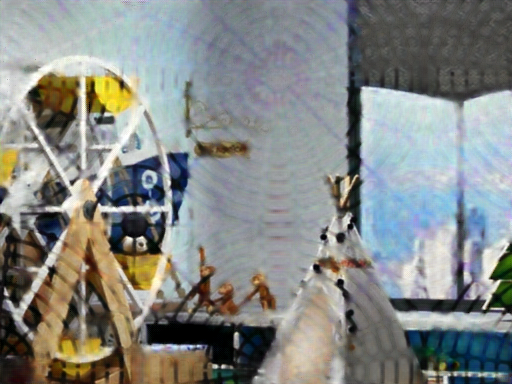} \hspace{-4mm} & 
        \includegraphics[width=.134\textwidth]{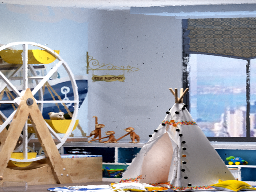} \hspace{-4mm} & 
        \includegraphics[width=.134\textwidth]{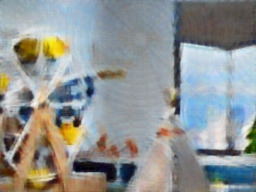} \hspace{-4mm} & 
        \includegraphics[width=.134\textwidth]{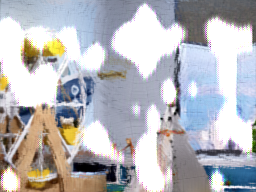} \hspace{-4mm} & 
        \includegraphics[width=.134\textwidth]{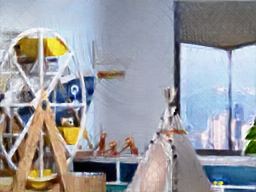} \hspace{-4mm} &
        \includegraphics[width=.134\textwidth]{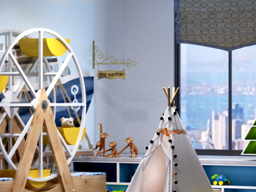} \hspace{-4mm} \\
        \includegraphics[width=.134\textwidth]{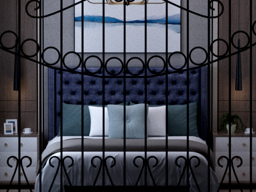} \hspace{-4mm} & 
        \includegraphics[width=.134\textwidth]{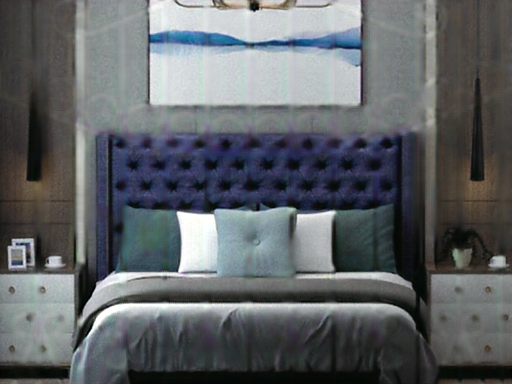} \hspace{-4mm} & 
        \includegraphics[width=.134\textwidth]{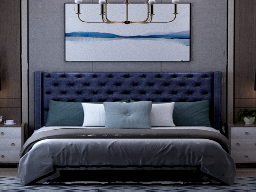} \hspace{-4mm} & 
        \includegraphics[width=.134\textwidth]{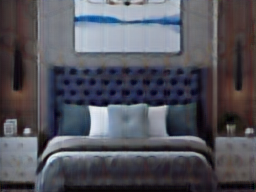} \hspace{-4mm} & 
        \includegraphics[width=.134\textwidth]{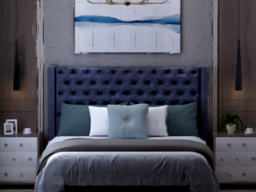} \hspace{-4mm} & 
        \includegraphics[width=.134\textwidth]{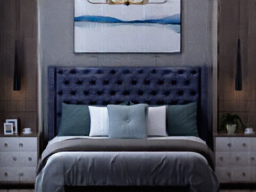} \hspace{-4mm} &
        \includegraphics[width=.134\textwidth]{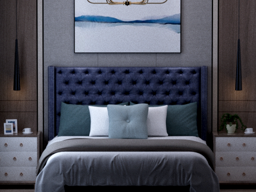} \hspace{-4mm} \\
        \includegraphics[width=.134\textwidth]{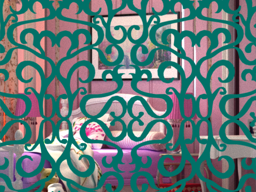} \hspace{-4mm} & 
        \includegraphics[width=.134\textwidth]{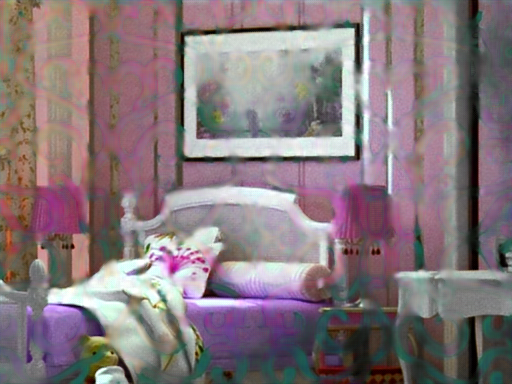} \hspace{-4mm} & 
        \includegraphics[width=.134\textwidth]{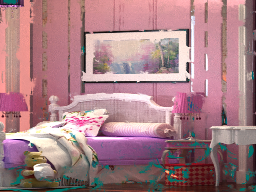} \hspace{-4mm} & 
        \includegraphics[width=.134\textwidth]{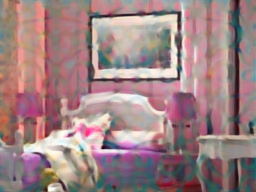} \hspace{-4mm} & 
        \includegraphics[width=.134\textwidth]{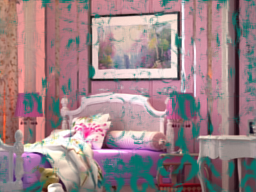} \hspace{-4mm} & 
        \includegraphics[width=.134\textwidth]{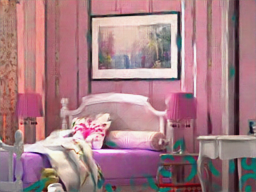} \hspace{-4mm} &
        \includegraphics[width=.134\textwidth]{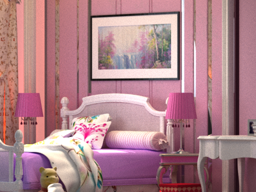} \hspace{-4mm} \\
        \includegraphics[width=.134\textwidth]{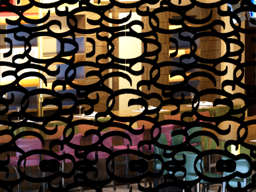} \hspace{-4mm} & 
        \includegraphics[width=.134\textwidth]{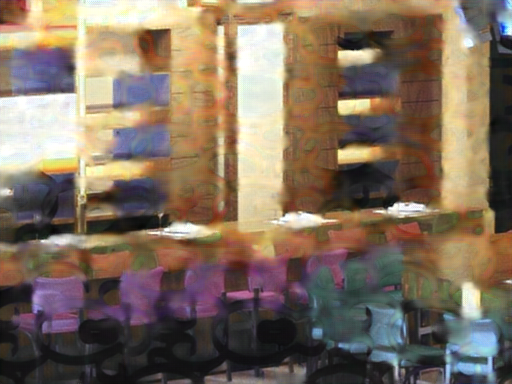} \hspace{-4mm} & 
        \includegraphics[width=.134\textwidth]{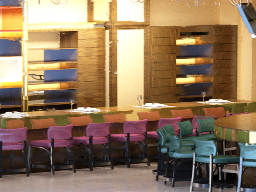} \hspace{-4mm} & 
        \includegraphics[width=.134\textwidth]{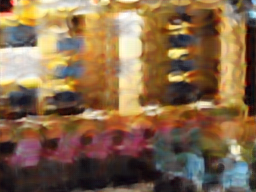} \hspace{-4mm} & 
        \includegraphics[width=.134\textwidth]{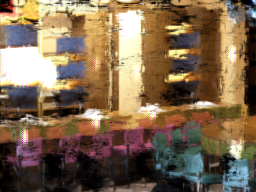} \hspace{-4mm} & 
        \includegraphics[width=.134\textwidth]{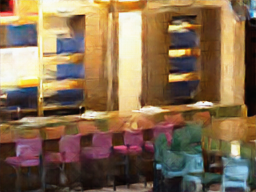} \hspace{-4mm} &
        \includegraphics[width=.134\textwidth]{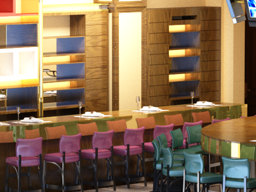} \hspace{-4mm} \\
        \includegraphics[width=.134\textwidth]{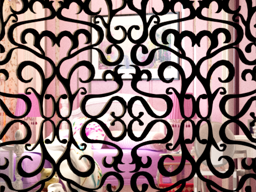} \hspace{-4mm} & 
        \includegraphics[width=.134\textwidth]{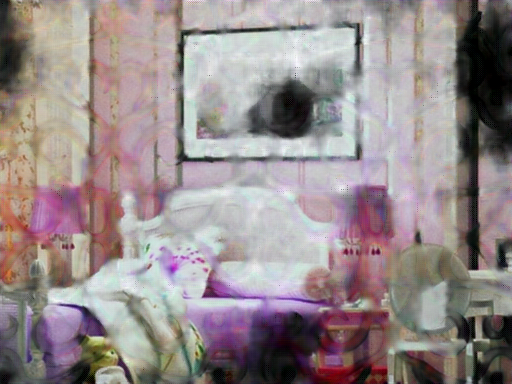} \hspace{-4mm} & 
        \includegraphics[width=.134\textwidth]{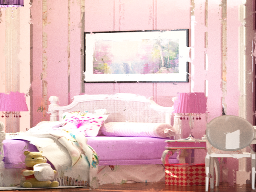} \hspace{-4mm} & 
        \includegraphics[width=.134\textwidth]{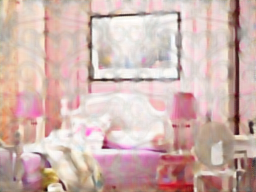} \hspace{-4mm} & 
        \includegraphics[width=.134\textwidth]{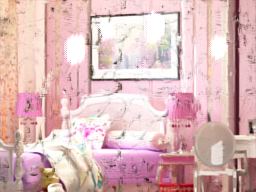} \hspace{-4mm} & 
        \includegraphics[width=.134\textwidth]{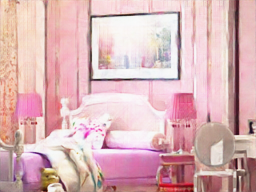} \hspace{-4mm} &
        \includegraphics[width=.134\textwidth]{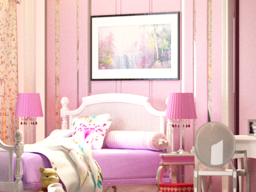} \hspace{-4mm} \\
        \includegraphics[width=.134\textwidth]{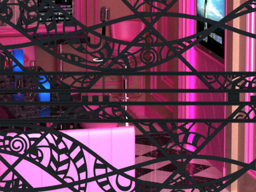} \hspace{-4mm} & 
        \includegraphics[width=.134\textwidth]{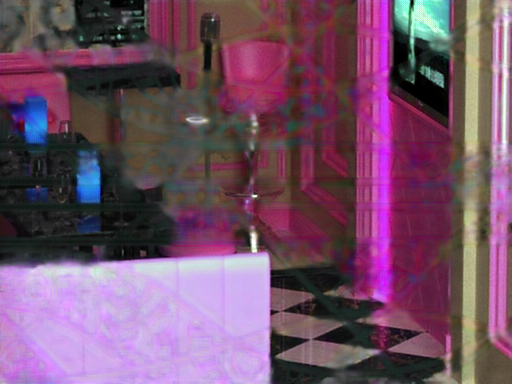} \hspace{-4mm} & 
        \includegraphics[width=.134\textwidth]{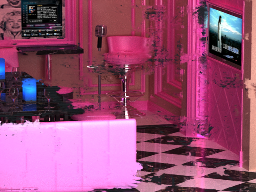} \hspace{-4mm} & 
        \includegraphics[width=.134\textwidth]{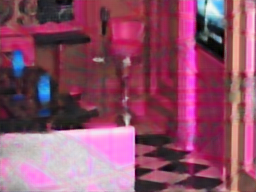} \hspace{-4mm} & 
        \includegraphics[width=.134\textwidth]{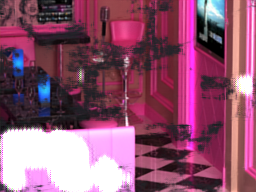} \hspace{-4mm} & 
        \includegraphics[width=.134\textwidth]{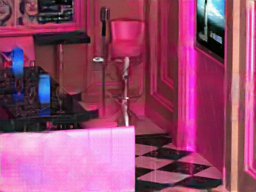} \hspace{-4mm} &
        \includegraphics[width=.134\textwidth]{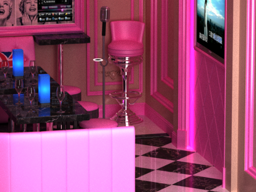} \hspace{-4mm} \\
        \includegraphics[width=.134\textwidth]{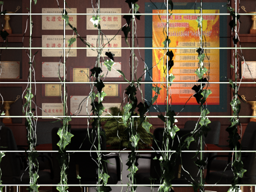} \hspace{-4mm} & 
        \includegraphics[width=.134\textwidth]{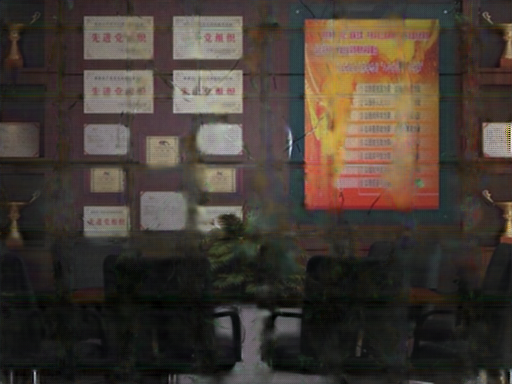} \hspace{-4mm} & 
        \includegraphics[width=.134\textwidth]{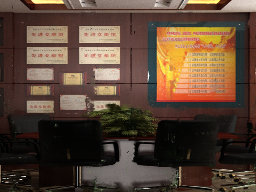} \hspace{-4mm} & 
        \includegraphics[width=.134\textwidth]{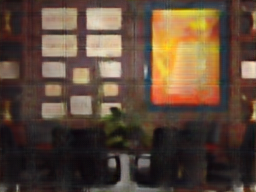} \hspace{-4mm} & 
        \includegraphics[width=.134\textwidth]{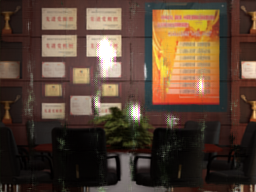} \hspace{-4mm} & 
        \includegraphics[width=.134\textwidth]{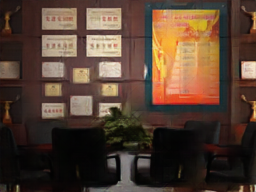} \hspace{-4mm} &
        \includegraphics[width=.134\textwidth]{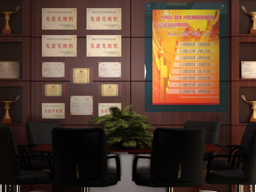} \hspace{-4mm} \\
        Input CV & DeOccNet & Zhang et al. & DeOccNet* & Zhang et al.* & Ours & GT
    \end{tabular}
    \caption{De-occlusion outputs on 9-syn dataset\cite{ijcai2021-180} using various LF-DeOcc methods, which is used for quantitative results. }
    \label{fig:9syn_outputs}
\end{figure*}

\textbf{Quantitative results.}
In this part, we provide the input, output and ground truth images used in quantitative results of the paper.
We calculate peak signal-to-noise ratio (PSNR) and structural similarity index measure (SSIM) for quantitative evaluation.
Figs.~\ref{fig:4syn_outputs} and \ref{fig:9syn_outputs} show the input CV images, ground truth occlusion-free CV images, and LF-DeOcc outputs.
Note that all scenes are synthetic sparse LF images, where most of the background objects are visible in other LF views.
As shown in qualitative results, our model generates the clear and accurate de-occlusion outputs with less occlusion artifacts.

\section{Experimental Results on Various Scenes}
\begin{figure*}[!h]
    \small
    \centering
    \begin{tabular}{cccccc}
        \includegraphics[width=.15\textwidth]{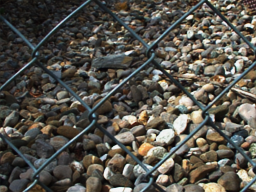} \hspace{-4mm} & 
        \includegraphics[width=.15\textwidth]{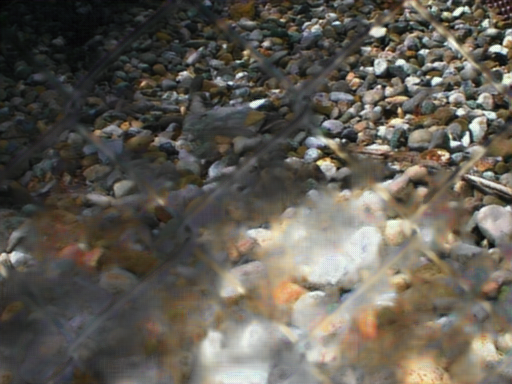} \hspace{-4mm} & 
        \includegraphics[width=.15\textwidth]{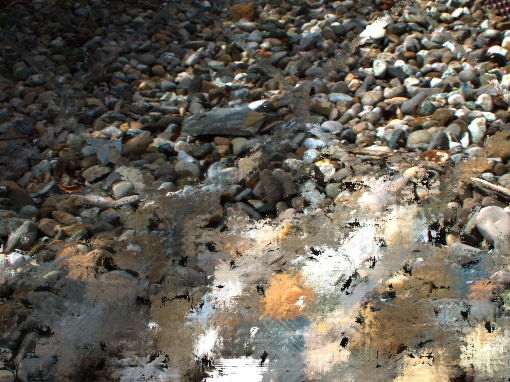} \hspace{-4mm} & 
        \includegraphics[width=.15\textwidth]{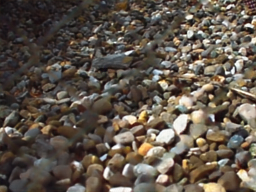} \hspace{-4mm} & 
        \includegraphics[width=.15\textwidth]{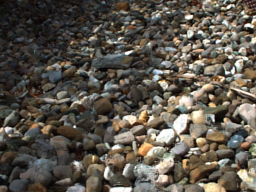} \hspace{-4mm} & 
        \includegraphics[width=.15\textwidth]{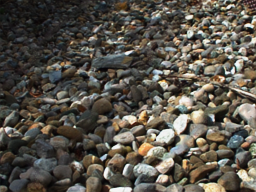} \hspace{-4mm} \\
        \includegraphics[width=.15\textwidth]{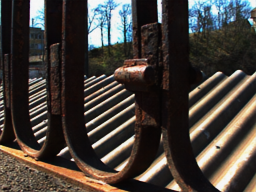} \hspace{-4mm} & 
        \includegraphics[width=.15\textwidth]{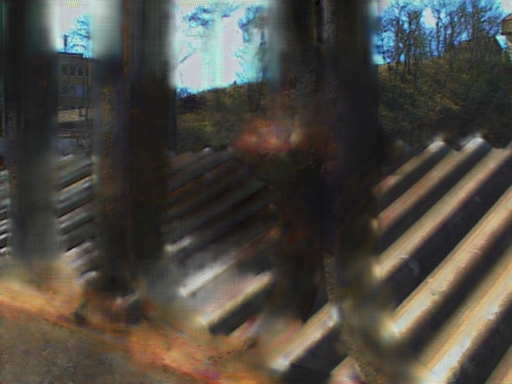} \hspace{-4mm} & 
        \includegraphics[width=.15\textwidth]{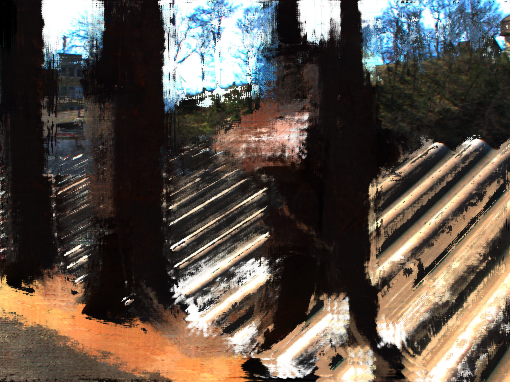} \hspace{-4mm} & 
        \includegraphics[width=.15\textwidth]{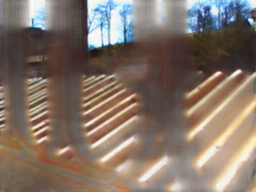} \hspace{-4mm} & 
        \includegraphics[width=.15\textwidth]{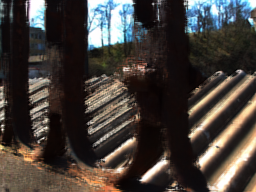} \hspace{-4mm} & 
        \includegraphics[width=.15\textwidth]{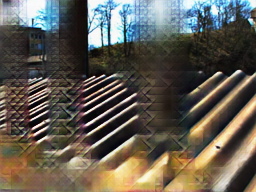} \hspace{-4mm} \\
        \includegraphics[width=.15\textwidth]{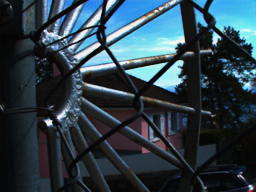} \hspace{-4mm} & 
        \includegraphics[width=.15\textwidth]{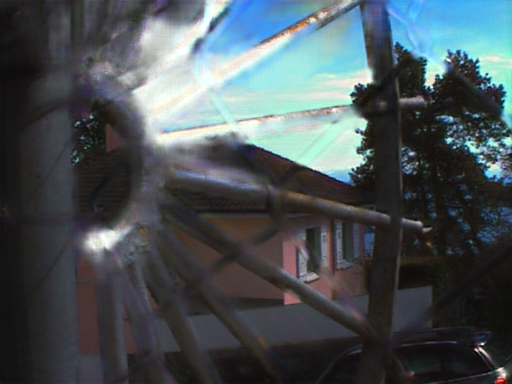} \hspace{-4mm} & 
        \includegraphics[width=.15\textwidth]{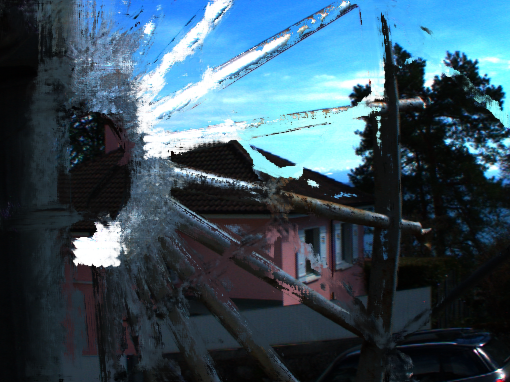} \hspace{-4mm} & 
        \includegraphics[width=.15\textwidth]{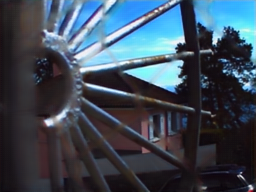} \hspace{-4mm} & 
        \includegraphics[width=.15\textwidth]{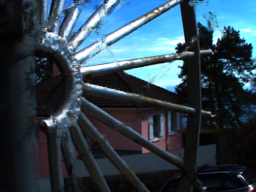} \hspace{-4mm} & 
        \includegraphics[width=.15\textwidth]{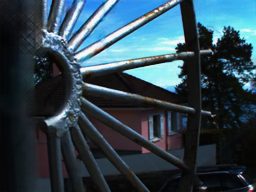} \hspace{-4mm} \\
        \includegraphics[width=.15\textwidth]{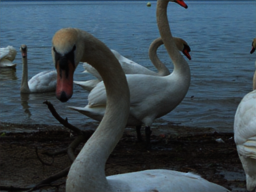} \hspace{-4mm} & 
        \includegraphics[width=.15\textwidth]{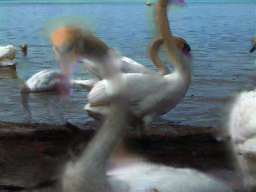} \hspace{-4mm} & 
        \includegraphics[width=.15\textwidth]{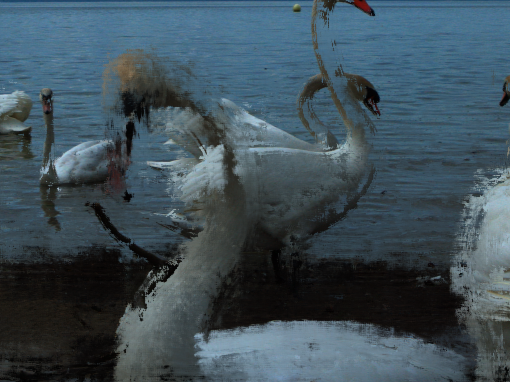} \hspace{-4mm} & 
        \includegraphics[width=.15\textwidth]{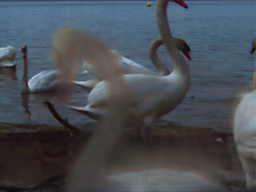} \hspace{-4mm} & 
        \includegraphics[width=.15\textwidth]{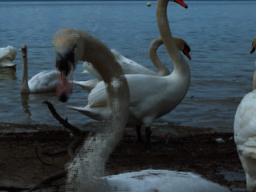} \hspace{-4mm} & 
        \includegraphics[width=.15\textwidth]{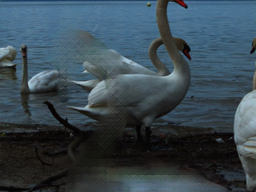} \hspace{-4mm} \\
        \includegraphics[width=.15\textwidth]{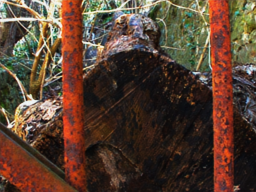} \hspace{-4mm} & 
        \includegraphics[width=.15\textwidth]{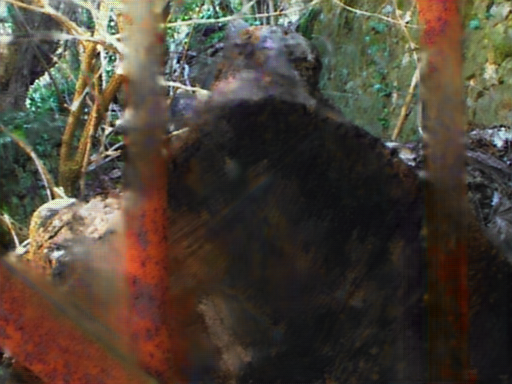} \hspace{-4mm} & 
        \includegraphics[width=.15\textwidth]{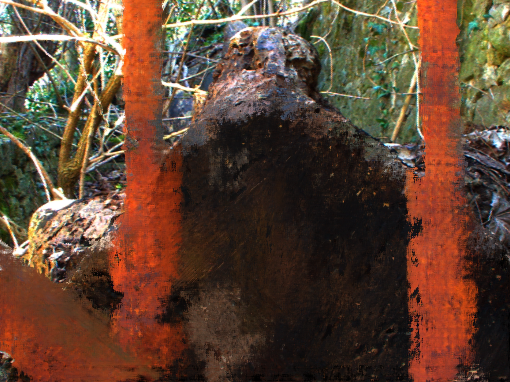} \hspace{-4mm} & 
        \includegraphics[width=.15\textwidth]{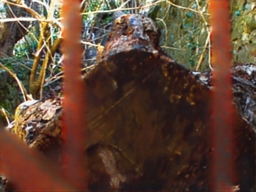} \hspace{-4mm} & 
        \includegraphics[width=.15\textwidth]{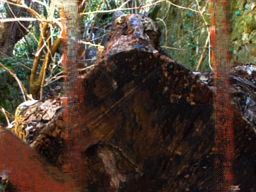} \hspace{-4mm} & 
        \includegraphics[width=.15\textwidth]{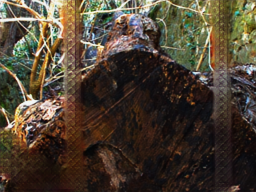} \hspace{-4mm} \\
        Input CV & DeOccNet & Zhang et al. & DeOccNet* & Zhang et al.* & Ours
    \end{tabular}
    \caption{De-occlusion outputs on various real-world occlusion scenes in EPFL-10 dataset\cite{epfl_dataset} which is a real-world dense LF dataset. }
    \label{fig:dense_output_epfl}
\end{figure*}

\begin{figure*}[!h]
    \small
    \centering
    \begin{tabular}{cccccc}
        \includegraphics[width=.15\textwidth]{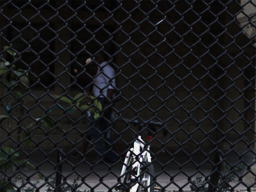} \hspace{-4mm} & 
        \includegraphics[width=.15\textwidth]{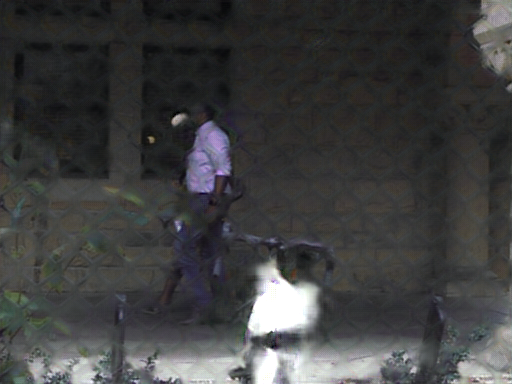} \hspace{-4mm} & 
        \includegraphics[width=.15\textwidth]{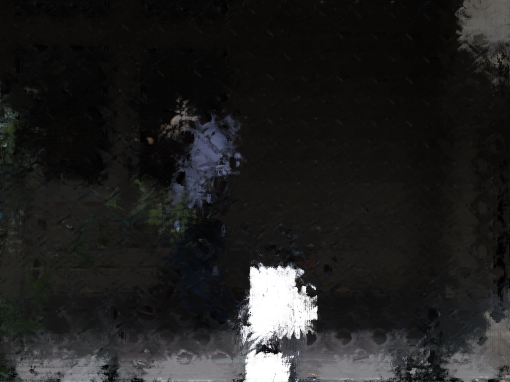} \hspace{-4mm} & 
        \includegraphics[width=.15\textwidth]{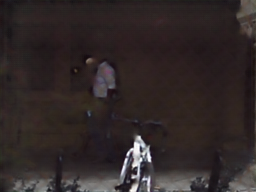} \hspace{-4mm} & 
        \includegraphics[width=.15\textwidth]{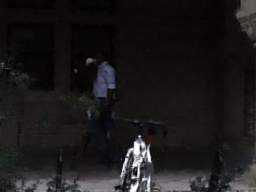} \hspace{-4mm} & 
        \includegraphics[width=.15\textwidth]{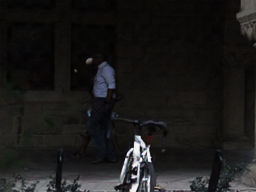} \hspace{-4mm} \\
        \includegraphics[width=.15\textwidth]{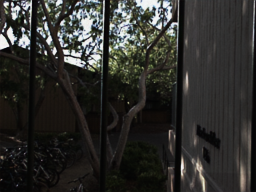} \hspace{-4mm} & 
        \includegraphics[width=.15\textwidth]{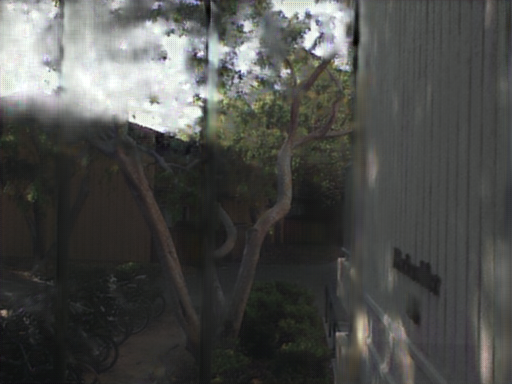} \hspace{-4mm} & 
        \includegraphics[width=.15\textwidth]{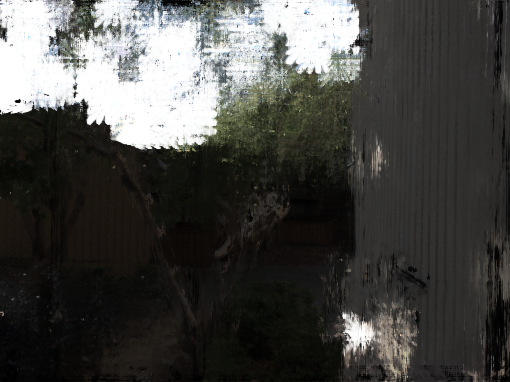} \hspace{-4mm} & 
        \includegraphics[width=.15\textwidth]{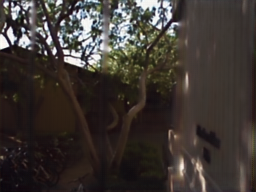} \hspace{-4mm} & 
        \includegraphics[width=.15\textwidth]{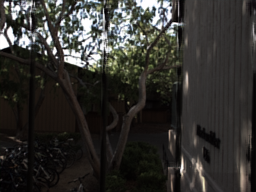} \hspace{-4mm} & 
        \includegraphics[width=.15\textwidth]{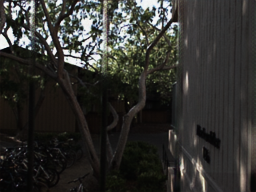} \hspace{-4mm} \\
        \includegraphics[width=.15\textwidth]{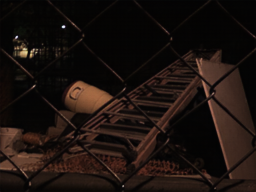} \hspace{-4mm} & 
        \includegraphics[width=.15\textwidth]{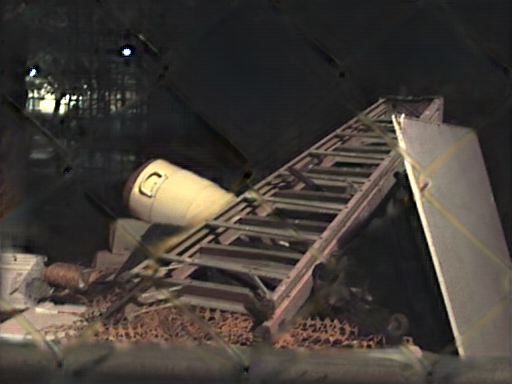} \hspace{-4mm} & 
        \includegraphics[width=.15\textwidth]{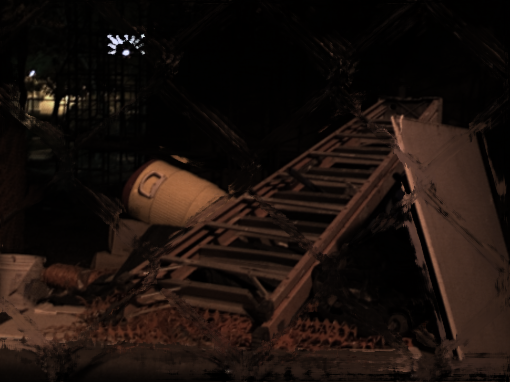} \hspace{-4mm} & 
        \includegraphics[width=.15\textwidth]{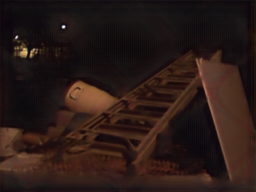} \hspace{-4mm} & 
        \includegraphics[width=.15\textwidth]{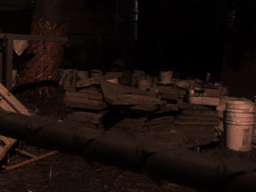} \hspace{-4mm} & 
        \includegraphics[width=.15\textwidth]{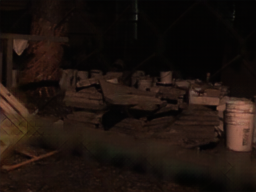} \hspace{-4mm} \\
        \includegraphics[width=.15\textwidth]{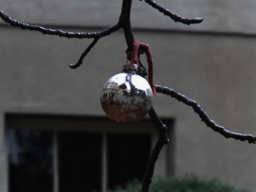} \hspace{-4mm} & 
        \includegraphics[width=.15\textwidth]{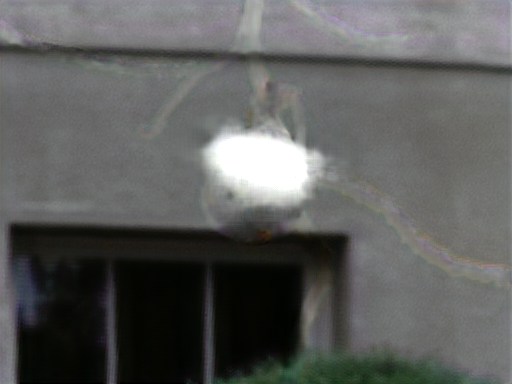} \hspace{-4mm} & 
        \includegraphics[width=.15\textwidth]{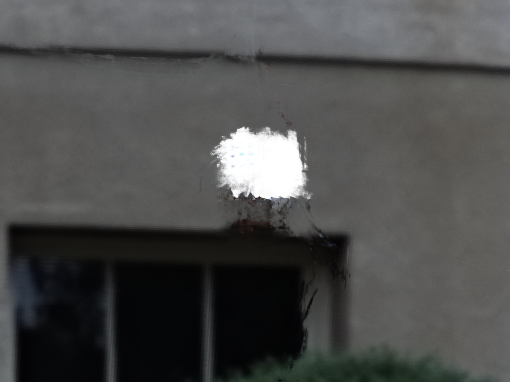} \hspace{-4mm} & 
        \includegraphics[width=.15\textwidth]{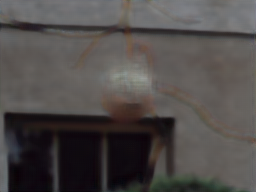} \hspace{-4mm} & 
        \includegraphics[width=.15\textwidth]{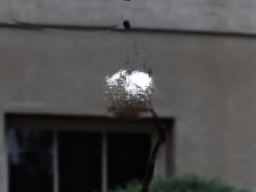} \hspace{-4mm} & 
        \includegraphics[width=.15\textwidth]{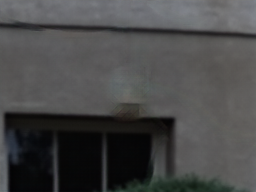} \hspace{-4mm} \\
        \includegraphics[width=.15\textwidth]{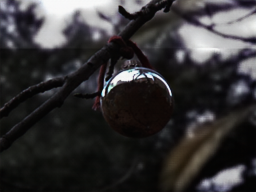} \hspace{-4mm} & 
        \includegraphics[width=.15\textwidth]{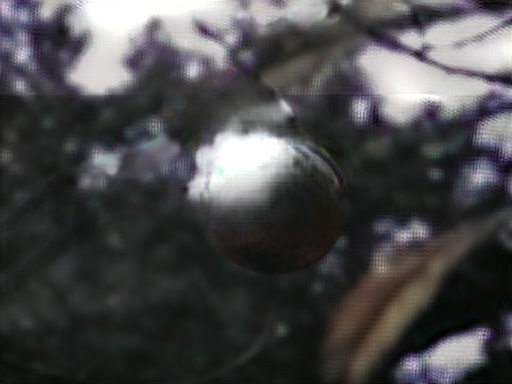} \hspace{-4mm} & 
        \includegraphics[width=.15\textwidth]{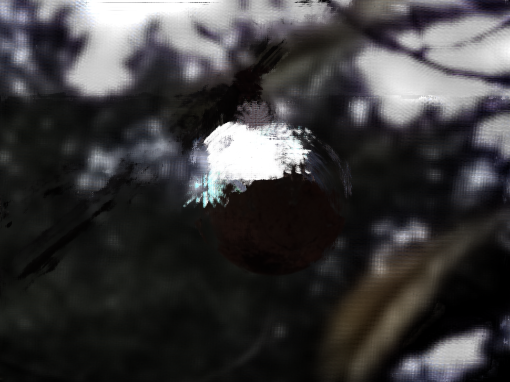} \hspace{-4mm} & 
        \includegraphics[width=.15\textwidth]{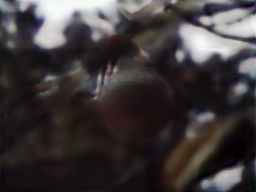} \hspace{-4mm} & 
        \includegraphics[width=.15\textwidth]{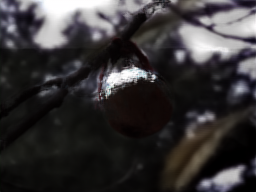} \hspace{-4mm} & 
        \includegraphics[width=.15\textwidth]{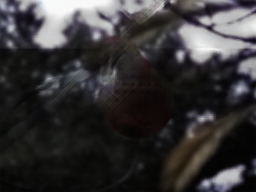} \hspace{-4mm} \\
        Input CV & DeOccNet & Zhang et al. & DeOccNet* & Zhang et al.* & Ours
    \end{tabular}
    \caption{De-occlusion outputs on various real-world occlusion scene in Stanford Lytro dataset\cite{stanford_dataset} which is a real-world dense LF dataset. }
    \label{fig:dense_output_stanford}
\end{figure*}

In this section, we provide various real-world LF-DeOcc outputs generated by existing and proposed LF-DeOcc methods.
Figs.~\ref{fig:dense_output_epfl} and \ref{fig:dense_output_stanford} shows the LF-DeOcc outputs of the dense LF images in EPFL-10 dataset~\cite{epfl_dataset} and Stanford Lytro dataset~\cite{stanford_dataset}.
Different from the existing LF-DeOcc methods, the proposed framework can remove the large occlusions in dense LF images, effectively preventing the artifacts from occlusions.
However, still some limitations can be seen as discussed in the main manuscript.
Since the inpainting knowledge is sub-optimal due to the relatively limited number of dataset compared to single RGB dataset (1418 scenes), the removed regions are sometimes unnatural if the occlusion is extremely large.
OI pre-trained on single RGB dataset may be effective, but the inpainting knowledge is catastrophically forgot during the LF-DeOcc training time. 
We expect some continual learning approach may solve this problem in exchange of memory, parameters, or training times.

\end{document}